\documentclass[preprint,12pt,authoryear]{elsarticle}

\usepackage{graphicx}%
\usepackage{multirow}%
\usepackage{amsmath,amssymb,amsfonts}%
\usepackage{amsthm}%
\usepackage{mathrsfs}%
\usepackage[title]{appendix}%
\usepackage{xcolor}%
\usepackage{textcomp}%
\usepackage{manyfoot}%
\usepackage{booktabs}%
\usepackage{algorithm}%
\usepackage{algorithmicx}%
\usepackage{algpseudocode}%
\usepackage{listings}%
\usepackage{verbatim}
\usepackage{amssymb}
\usepackage{amsmath}
\usepackage{ragged2e}
\usepackage{url}
\journal{Computers \& Graphics}

\begin{document}

\begin{frontmatter}

\title {VITON-DRR: Details Retention Virtual Try-on via Non-rigid Registration}

\author[li]{Ben~Li}
\author[min,co]{Minqi~Li}
\author[ren]{Jie~Ren}
\author[zh]{Kaibing~Zhang}
\affiliation[li]{organization={Xi’an Polytechnic University},
	addressline={210411043@stu.xpu.edu.cn}, 
	city={Xi’an},
	postcode={710048}, \\
	state={Shannxi},
	country={China}}
\affiliation[min]{organization={Xi’an Polytechnic University},
	addressline={minqi.li@xpu.edu.cn}, 
	city={Xi’an},
	postcode={710048}, 
	state={Shannxi},
	country={China}}
\affiliation[ren]{organization={Xi’an Polytechnic University},
	addressline={renjie@xpu.edu.cn}, 
	city={Xi’an},
	postcode={710048}, 
	state={Shannxi},
	country={China}}
\affiliation[zh]{organization={Xi’an Polytechnic University},
	addressline={zhangkaibing@xpu.edu.cn}, 
	city={Xi’an},
	postcode={710048}, 
	state={Shannxi},
	country={China}}
\affiliation[co]{Corresponding author}

\begin{abstract}
	
Image-based virtual try-on aims to fit a target garment to a specific person image and has attracted extensive research attention because of its huge application potential in the e-commerce and fashion industries. To generate high-quality try-on results,  accurately warping the clothing item to fit the human body plays a significant role, as slight misalignment may lead to unrealistic artifacts in the fitting image. Most existing methods warp the clothing by feature matching and thin-plate spline (TPS). However, it often fails to preserve clothing details  due to self-occlusion, severe misalignment between poses, etc. To address these challenges, this paper proposes a detail retention virtual try-on method via accurate non-rigid registration (VITON-DRR) for diverse human poses. Specifically, we reconstruct a human semantic segmentation using a dual-pyramid-structured feature extractor. Then, a novel Deformation Module is designed for extracting the cloth key points and warping them through an accurate non-rigid registration algorithm. Finally, the Image Synthesis Module is designed to synthesize the deformed garment image and generate the human pose information adaptively. {Compared with} traditional methods, the proposed VITON-DRR can make the deformation of fitting images more accurate and retain more garment details. The experimental results demonstrate that the proposed method performs better than state-of-the-art methods. Our code is publicly available at https://github.com/minqili/VITON-DRR-main.

\end{abstract}

\begin{graphicalabstract}
\includegraphics[width=1\linewidth]{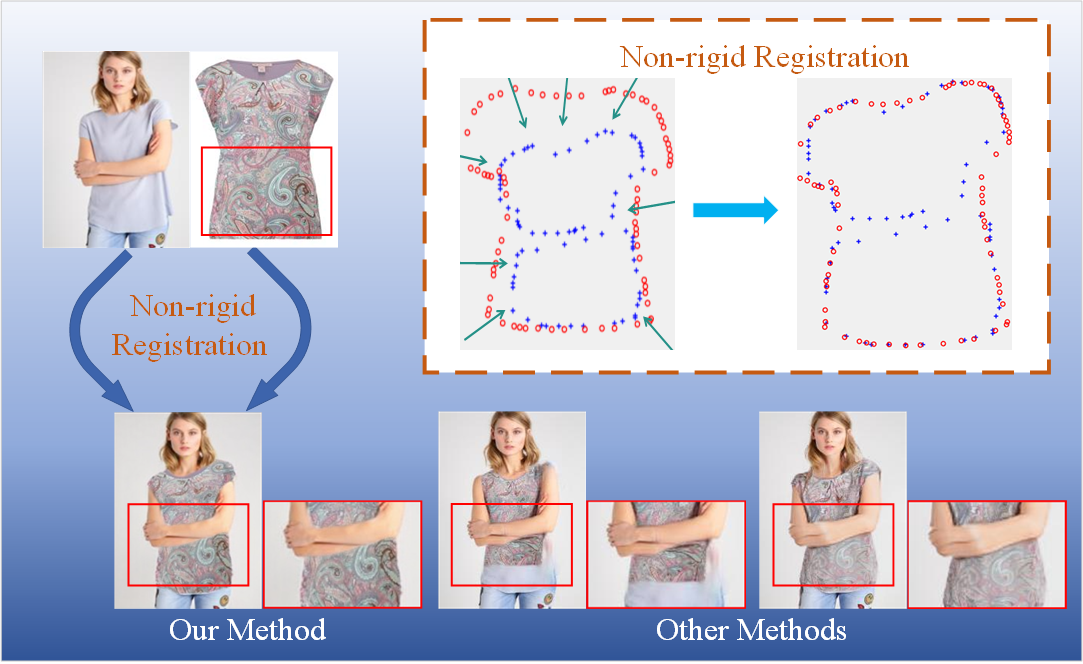}
\end{graphicalabstract}


\end{frontmatter}


\section{Introduction}

With the growing demand for online clothing shopping, the virtual try-on system has gained significant attention. It can anticipate the wearing effect of clothing, increase the diversity of people's clothing choices, and improve the online shopping experience. According to the different effects of virtual fitting, virtual try-on methods can be divided into 3D virtual try-on methods and 2D virtual try-on methods. Among them, the 3D virtual try-on method can show a 3D effect, but it needs to establish a 3D model and carry out additional 3D measurement, which requires high computing power and hardware \cite{chen2024gaussianvton, pons2017clothcap, bhatnagar2019multi, patel2020tailornet, brouet2012design}.
In contrast, 2D image-based methods \cite{han2018viton, wang2018toward, yu2019vtnfp, yang2020towards, ishikawa2022image, zhang2024gic, chen2024cs, adhikari2023vton} are fast to implement and have low hardware requirements. The 2D virtual try-on method not only better maintains the real details of the garment images, but also makes it easy to promote the application. This makes it the mainstream of current research.

2D image-based virtual try-on methods aim to transfer the target garment to the corresponding area of the reference person naturally and accurately. In order to preserve the body posture information and clothing details information as much as possible, virtual try-on methods should meet the following requirements \cite{han2018viton, wang2018toward}: 
1). the original body parts, shape, and posture should be well preserved;
2). the deformation of the clothing image should be natural and as close as possible to the fitting area;
3). the patterns and details of the clothes after deformation should be well preserved; 
4). the image should be rendered correctly even when body parts are occluded (e.g., hands should be rendered correctly when trying on short sleeves while wearing long sleeves).

To meet the above requirements and achieve high-quality virtual try-on, most of the previous methods involved two processes:
a) deforming the target garment according to the given human body shape;
b) fusing the deformed garment image with the human body image while preserving the texture details. This suggests that the accuracy of garment image deformation is crucial to obtaining high-quality try-on results. However, how to perform accurate garment deformation remains a challenging problem, and most existing methods attempt to solve it by relying on Thin Plate Spline (TPS) \cite{bookstein1989principal} or appearance flow \cite{huang2022towards}.

\begin{figure}[htb]
	\begin{center}
		\includegraphics[width=0.8\linewidth]{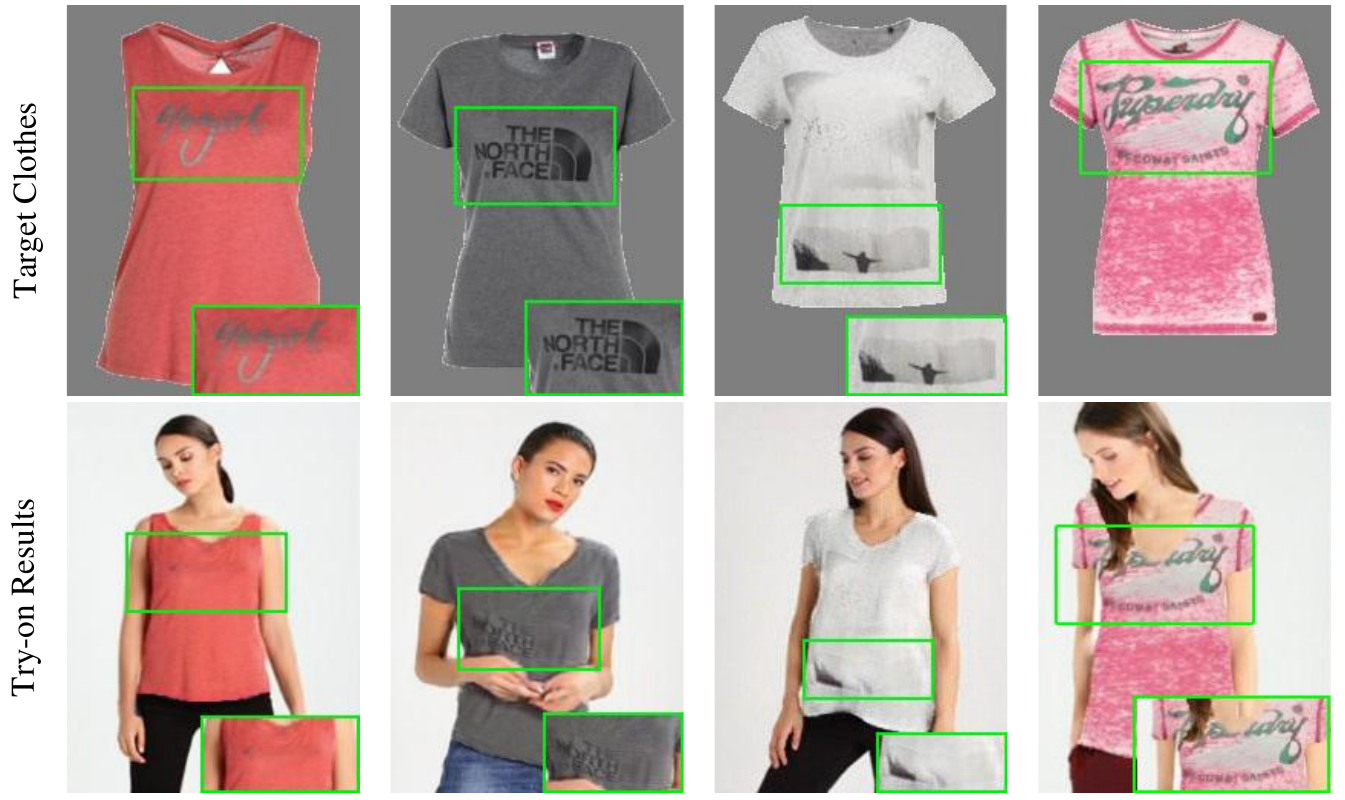}
	\end{center}
	\caption{Example of missing clothing details in virtual try-on results. The fitting results in the above figure lost the pattern details of the clothes due to unreasonable deformation and fusion of the clothes.} 
	\label{fig_1}
\end{figure}

\begin{figure}[htb]
	\begin{center}
		\includegraphics[width=0.8\linewidth]{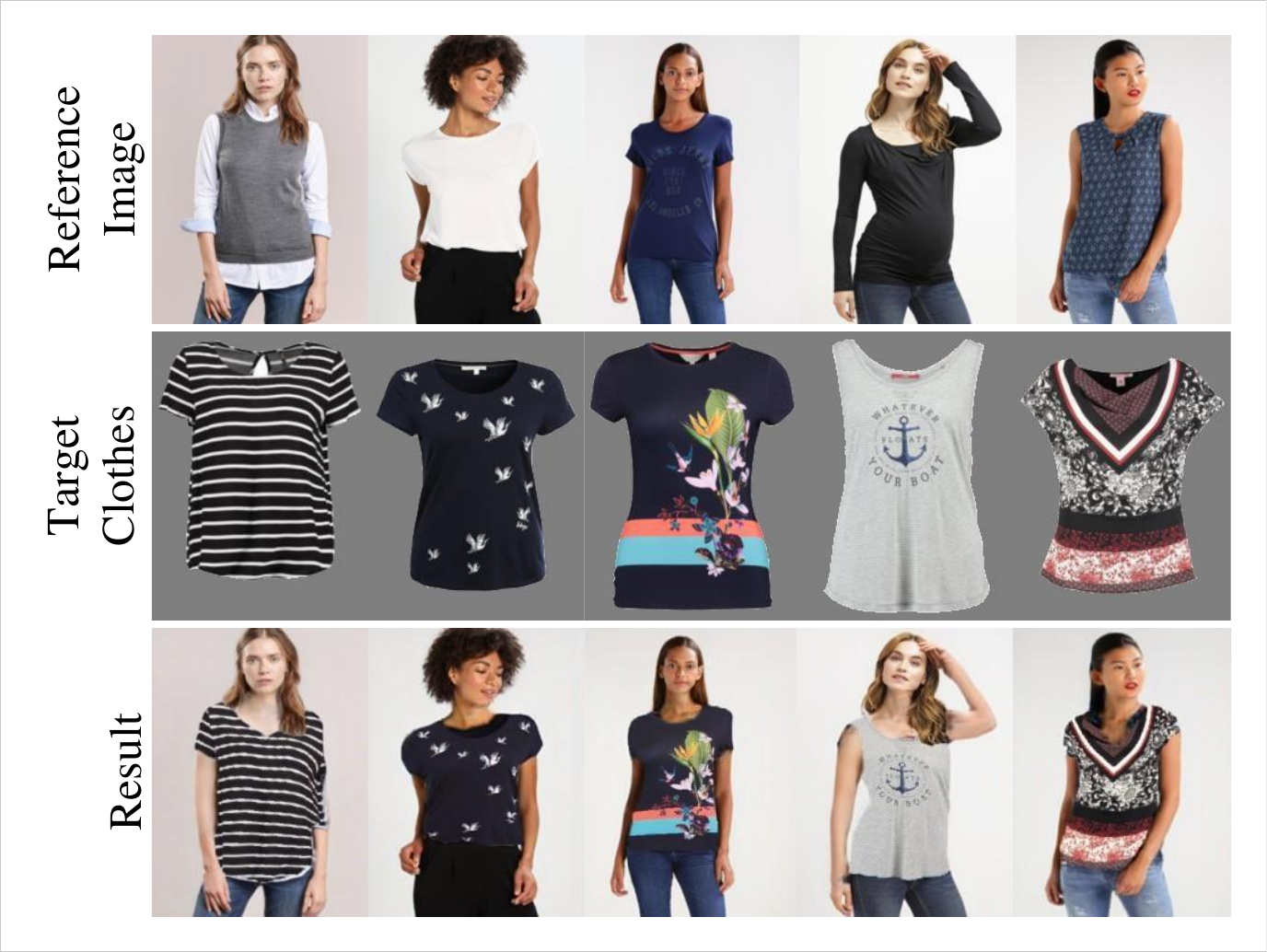}
	\end{center}
	\caption{Examples of virtual try-on results synthesized by our method. Given a reference human body image and a product clothing image, our method can synthesize a high-quality virtual try-on result while preserving garment details.}
	\label{fig_2}
\end{figure}

The TPS-based method first extracts shape features and matching regions to obtain the transformation parameters. Then, TPS is used to warp the target image to the corresponding reference image region. Due to the limited warping degrees of freedom based on TPS transformation, for large or complex deformations, although the image deformation results obtained by TPS are smooth, it is difficult to retain the detailed texture \cite{schaefer2006image, he2022style}, for example, when different regions of a garment undergo large and different deformations.

Among the flow estimation methods, most existing methods construct optical flow maps based on local feature correspondences \cite{he2022style, lee2022high}. For example, the appearance flow \cite{ge2021parser} indicates which pixels in the source can synthesize the target with reference to 2D coordinate vectors. ClothFlow \cite{han2019clothflow} estimates the optical flow maps of the clothes and the desired clothing regions through a cascaded appearance flow estimation network, which progressively deforms the source image features and refines the flow prediction. To reduce undesired output artifacts, ZFlow \cite{chopra2021zflow} combines gate-like appearance flows with dense geometric priors for human pose transfer.
However, flow estimation methods are limited to areas of clothing with small non-rigid deformation, because they assume that the corresponding area in the person image is in the same local area as the garment. However, when there is a large deformation between the garment and the corresponding body parts, this assumption is unrealistic \cite{bai2022single, he2022style}. As shown in Fig. \ref{fig_1}, when the clothes undergo large deformation, the clothing details are missing in the virtual fitting results, which is a common challenge in most existing virtual fitting methods.

To address the above challenges, we propose a novel Details Retention virtual try-on method, called VITON-DRR. VITON-DRR is an image-based method designed to synthesize realistic images of a person wearing a target garment while preserving clothing details through human semantics generation and accurate non-rigid registration. As shown in Fig. \ref{fig_2}, our method can maintain fine-grained clothing details across different human poses and achieve superior try-on effect. The method proposed in this paper abandons the conventional methods that rely on deformation networks to estimate deformation parameters and employ TPS to deform images. Instead, we improve the deformation estimation through three steps: a) extracting the salient points on the edge of the target garment and the edge of the target region to better represent the corresponding shapes; b) aligning the two sets of feature points using a non-rigid registration method to obtain the accurate garment deformation parameters; c) using moving least squares (MLS) to deform the image to ensure the rationality of the image deformation and the integrity of details. This framework addresses key limitations in existing methods by: Enhancing geometric alignment through semantically guided point matching; Preserving  details (e.g., textures, patterns) via non-rigid registration.

The main contributions of this paper are summarized as follows:
\begin{itemize}
	\item[-] We propose a novel virtual try-on network called VITON-DRR that solves the problem of garment detail retention through accurate non-rigid registration. This method can adapt to  various deformations of human posture and preserve garment details in the deformed fitting region.
	\item[-] We design a novel Deformation Module to calculate the matching by extracting the key point features of clothing areas and estimate the deformation by a Moving Least Squares (MLS) algorithm. In this way, the correspondences between the target area mask and the clothing mask can be calculated precisely under conditions of large deformation and self-occlusion. Moreover, compared with the traditional cloth deformation strategies in virtual try-on methods, our method makes the image warping smoother and more natural.
	\item[-] We conduct extensive quantitative and qualitative comparisons to verify that the proposed method yields more impressive results in preserving clothing detail compared to other state-of-the-art competitors in various deformation situations.
\end{itemize}

\section{Related Work}
In virtual try-on methods, accurate garment deformation and fusion with the figure image are two key tasks for generating high-quality virtual try-on results. To accomplish these tasks, the joint application of multiple techniques such as image segmentation, image synthesis, and image registration are required.

\subsection{ Image Segmentation and Synthesis }
Recently, Generative Adversarial Network (GAN) has become one of the most popular deep learning models for image generation \cite{stein2024exposing, takida2024san}, and has demonstrated promising results in tasks such as image segmentation \cite{zhu2020semantically, usman2024brain}, texture synthesis \cite{chen2024integrating, zhou2023neural}, image super-resolution \cite{wang2023simultaneous, bouchard2023resolution}, object detection \cite{bosquet2023full}, video \cite{cherian2024generative, tulyakov2018mocogan}, \cite{wang2021generative}, \emph{etc.}.

In the field of virtual try-on, GAN based methods are often used for body and clothing-agnostic semantic segmentation or try-on result synthesis. For example, Jetchev \emph{et al}. \cite{jetchev2017conditional} treat virtual try-on as a style transfer task. They introduced CGAN \cite{mirza2014conditional} in CycleGAN \cite{zhu2017unpaired} to solve the garment item swap problem. In  \cite{yu2019vtnfp, yang2020towards}, GANs are  used for the generation of human semantic segmentation. In \cite{han2018viton, wang2018toward, dong2019towards},  GANs are designed in the coarse synthesis stage of virtual try-on. Inspired by these impressive GAN-based results, we also apply adversarial losses to develop the semantic segmentation and virtual try-on synthesis with GANs.

\subsection{Image Registration}
Image registration is an important fundamental research topic in computer vision. It can be described as a point set registration problem by performing feature extraction on the target and reference images. Point cloud registration aims to find the transform relationship between two point sets and perform operations such as translation and rotation on the point sets to make them correspond to each other. Depending on the underlying transformation model, point cloud registration can be broadly classified into rigid or non-rigid. Iterative Closest Point (ICP) and its variants \cite{marin2024nicp, lin2024icp} have achieved good results in rigid point cloud registration, but have not performed well in non-rigid object registration. Non-rigid registration requires estimating of a set of local transformations, which is well applied in the fields of medical image analysis, 3D object pose estimation, VR, target tracking, etc., and has been extensively studied in the past decades \cite{du20133d, liu2016registration, delavari2019accurate}. In virtual try-on systems, the deformation between the desired garment and the corresponding region of the reference person is usually non-rigid. Most of the existing virtual try-on methods use image alignment techniques such as TPS to solve the deformation problem \cite{han2018viton, wang2018toward, yu2019vtnfp, yang2020towards}. For example, VITON computes the transformation mapping through the shape context TPS warping \cite{bookstein1989principal}. \cite{wang2018toward} designed a convolutional neural network (CNN) to estimate the TPS transformations. \cite{dong2019towards} warped the feature maps from a bottleneck layer of Warp-GAN by using both affine and TPS. However, due to the deficiency of warping methods, i.e., limited degrees of freedom in TPS, these methods cannot model highly non-rigid deformations.

\subsection{Image-based Virtual Try-On}

Image-based virtual try-on aims to synthesize a realistic image of a model wearing a garment image (the target garment)  given a pair of 2D garment and person images. Earlier work by Jetchev \emph{et al}. \cite{jetchev2017conditional} demonstrated the concept and preliminary results  of using CAGAN to swap human garments, opening up new possibilities for image processing for fashion purposes. However, this approach does not take into account the effects of changes in body posture. The later approaches take into account different body postures by using two additional modules: a) the Semantic Generation Module for rearranging the semantic segmentation of the human body to ensure accurate subsequent warping of the garment to the human body; and b) the Clothing Warping Module for warping the clothes naturally onto a human body with various postures. Finally, the photo-realistic try-on images are synthesized by utilizing intermediate representations generated in the previous stage (e.g., warped clothes and segmentation maps). For example, VITON \cite{han2018viton} adopts a coarse-to-fine strategy to achieve 2D virtual try-on with different body poses. Due to the imperfect shape-context matching between clothing and body shape, and the inferior appearance merging strategy, the performances are moderate. To improve the accuracy of image deformation, CP-VITON \cite{wang2018toward} introduces a geometric matching module based on VITON to learn the deformation parameters of TPS. VTNFP \cite{yu2019vtnfp} utilizes an additional semantic segmentation module to reconstruct the human body semantic segmentation map and ensures adaptive retention of human body features. Meanwhile, regarding image deformation, VTNFP learns the deformation parameters of TPS using a double-cascade CNN. Similarly, ACGPN \cite{yang2020towards} uses additional semantic segmentation modules. The difference is that this method uses a spatial transformation network (STN) with second-order difference constraints to learn the deformation parameters of TPS. In addition, ClothFlow \cite{han2019clothflow}, PF-AFN \cite{ge2021parser}, Flow-Style \cite{he2022style} and DAFlow \cite{bai2022single} use Appearance Flow \cite{ge2021parser} based methods to predict garment deformation. However, the performance of these methods in terms of detail retention is still not entirely satisfactory.


\begin{figure}[th!]
	\centering
	\begin{center}
		\includegraphics[width=1\linewidth]{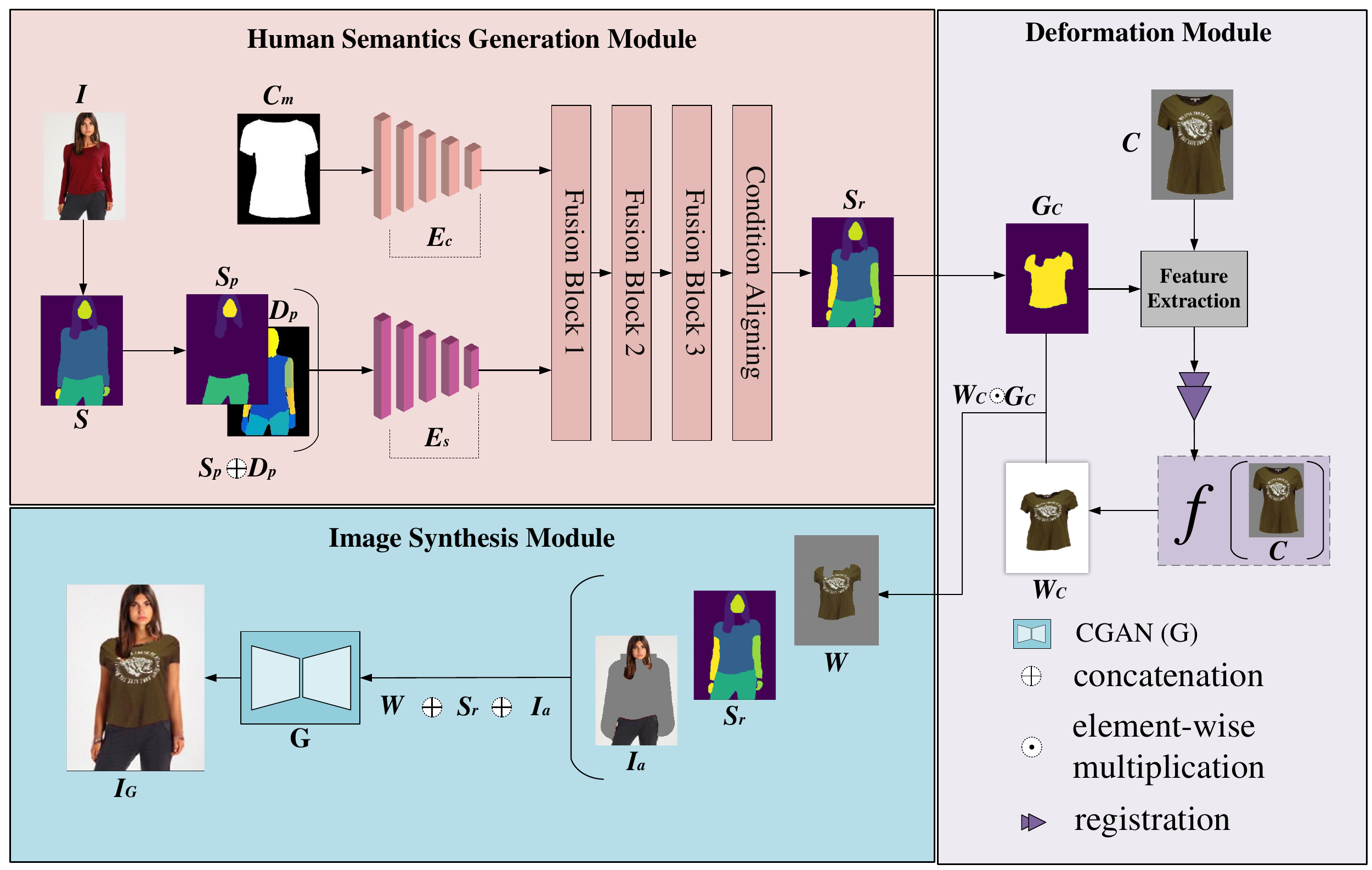}
	\end{center}
	\caption{An overview of our VITON-DRR, containing three main modules. 1. (a) Human Semantics Generation Module predicts the human semantic segmentation $S_r$  given garment segmentation inputs $(C_m, S_p, D_p)$, where $S_p$ is the clothing-agnostic semantic segmentation from the human semantic segmentation $S$, $D_p$ is the pose mapping, and $C_m$ is the clothing mask from the garment $C$. (b) Deformation Module warps $C$ through the feature point registration function $f$ with the target segmentation template $G_C$ to obtain the deformed garment $W_C$, where $G_C$ is the garment region obtained from $S_r$. (c) Image Synthesis Module synthesizes the final virtual try-on result $G_I$ using $W$, $S_r$, and $I_a$, where $I_a$ is the clothing-agnostic human image.}
	\label{fig_3}
\end{figure}

\section{Proposed Method}
The target of the proposed VITON-DRR is, given a reference person image $I \in \mathbf{R}^{H \times W \times 3}$ (H and W represent the length and width of the image, respectively) and a target garment $C \in \mathbf{R}^{H \times W \times 3}$, to naturally and accurately deform $C$ to the target area and ensure that the body posture information and the clothing detail information are preserved as much as possible. To this end, VITON-DRR consists of three modules, as shown in Fig. \ref{fig_3}.

Firstly, a Human Semantics Generation Module (HSGM) is designed to rearrange the semantic segmentation of the human body. In this way, the human posture features are preserved during the virtual try-on, and the matching area of the target cloth is marked (section \ref{HSGM}). Then, a Deformation Module (DM) utilizes a non-rigid registration method to accurately match the edge points of the target garment with the edge points of the target region. Next, the deformation parameters are calculated and further applied to garment warping (section \ref{DM}). Finally, a Image Synthesis Module (ISM) is designed to synthesize the final virtual try-on result (section \ref{ISM}).

\subsection{Human Semantics Generation Module (HSGM)} \label{HSGM}
To ensure natural connections between the body parts and the 
target garment shapes during the virtual try-on process, we utilize a conditional generative adversarial network (CGAN) to reconstruct the human semantic segmentation map. The predicted semantic segmentation map can more accurately locate the filled regions of the target garment and facilitate accurate deformation of the garment image. In addition, using the reconstructed semantic segmentation map, the body parts can be retained or covered adaptively to ensure the rationality of the virtual fitting results and the integrity of the human posture.

In HSGM, the conditional generator consists of two feature encoders ($E_c$ and $E_s$), three feature fusion blocks, and a conditional alignment block. It takes the target clothing mask $C_m$ and human semantic segmentation representation as input. The human semantic segmentation consists of a garment-independent human semantic segmentation $S_p$ and a pose map $D_p$ \cite{guler2018densepose}). The feature encoder extracts features from the given input and then fuses them in the feature fusion blocks. Finally, the prediction results are mapped out through the conditional alignment block.  

We utilize ternary input $(S_p, D_p, C_m)$ to train HSGM aimed at generating a new human semantic segmentation $S_r$. Specifically, $E_c$ and $E_s$ are employed to extract the five-layer pyramid features ${{\{E_{ck}\}}^4_{k=0}}$ and ${{\{E_{sl}\}}^4_{l=0}}$ of $(S_p, D_p)$ and $C_m$, respectively. Then, the Fusion Blocks merge the feature maps of these two components. Finally, these features are passed through the conditional alignment layer to output the reconstructed human semantic segmentation mask $S_r$.

The input to the Fusion Block consists of two information streams that exchange information with each other. After processing by the Fusion Blocks, the output is provided by the Condition Aligning Block \cite{wang2018high} to predict the human semantic segmentation.

The difference between the predicted segmentation map $S_r$ and the real segmentation map $S$ is evaluated using Pixel-wise Cross-Entropy Loss $L_{CE}$ and the CGAN Loss $L_{CGAN}$. The overall loss function for HSGM is defined as:

\begin{equation}
	L_{HSGM} = L_{CGAN} + \lambda_1 L_{CE}\textcolor{blue}{,}
\end{equation}
where $\lambda_1$ is a hyper-parameter that weighs the relative importance between the two loss functions, $L_{CGAN}$ and $L_{CE}$.

\subsection{Deformation Module (DM)} \label{DM}
One challenge with virtual try-on is how to accurately warp the target clothing image to fit the human pose. In previous works, the commonly used methods include TPS and flow estimation for geometric transformation \cite{han2018viton, wang2018toward, yu2019vtnfp, he2022style, bai2022single}. In general, these methods have smooth and reasonable image deformation results. 
However, when the local changes are drastic and the overall deformation is not large, even if the clothing image can be matched, the internal parts of the images (e.g., text and logos on the garments) may still appear unreasonable distortion or lead to blurred results.

\begin{figure}[htb]
	\centering
	\begin{center}
		\includegraphics[width=0.65\linewidth]{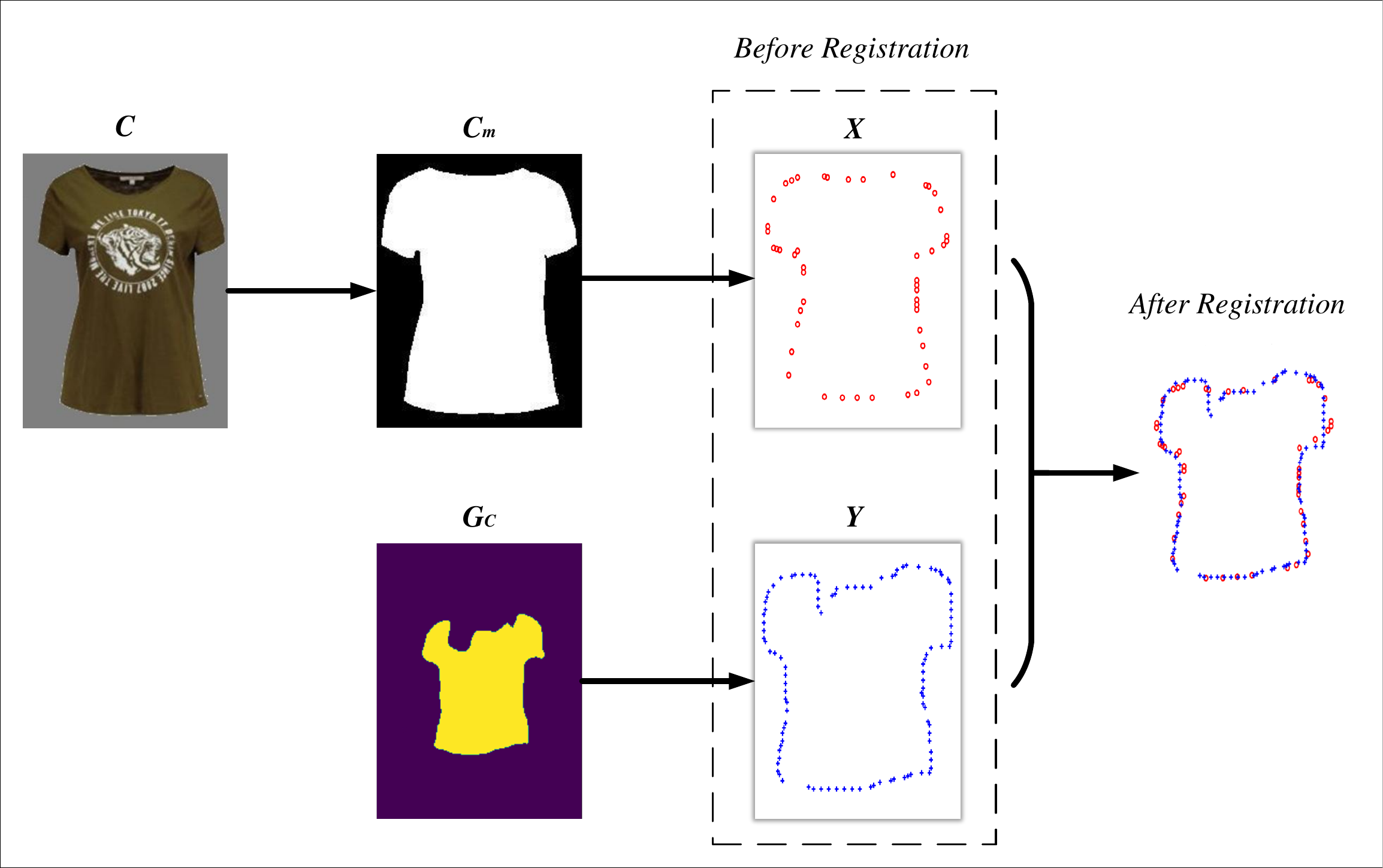}
	\end{center}
	\caption{Example of the clothes image registration. Given the clothes mask $C_M$ and the target area mask $G_C$, the edge point clouds $X$ and $Y$ are extracted, respectively. We register $X$ and $Y$ using a non-rigid point cloud registration method. The deformation parameters of $C$ are determined by the spatial motion of the point cloud $X$ during the registration process.}
	\label{fig_4}
\end{figure}

To solve the above problem, we introduce an accurate non-rigid registration method \cite{li2020fast} to match the edge keypoints of the target area mask with the edge keypoints of the target clothing mask, as shown in Fig. \ref{fig_4}. The trajectories of the points are recorded and used to calculate the garment deformation parameters.

Suppose we have two point sets $\mathbf{X}$ and $\mathbf{Y}$ in $R^2$, consisting of points $\{\mathbf{x}_i, i=1, 2, ...,m\}$ and $\{\mathbf{y}_j, j=1, 2, ..., n\}$, respectively. The point $\mathbf{x}_i$ is considered to be mapped to a new location $f(\mathbf{x}_i)$, where $f$ denotes the non-rigid transformation. The goal of non-rigid registration can be expressed as minimizing the following binary learning assignment-least squares energy function \cite{chui2003new}:
\begin{equation}\label{eq:Q}
	\mathop{\min}_f E(f)= \mathop{\min}_f \sum_{j=1}^n \sum_{i=1}^m p_{ji}\| \mathbf{y}_j-f(\mathbf{x}_i) \|^2+\lambda_2 \| f \|^2,
\end{equation}
where $p_{ji}$ is the probability for $\mathbf{y}_j$ corresponding to $\mathbf{x}_i$, and $\lambda_2$ is a positive constant for regularization. In our application, $\mathbf{y}$ and $\mathbf{x}$ are the edge contour points of target and reference clothes, respectively. Typically, EM algorithm is used to compute the correspondence $p_{ji}$ and deformation function $f$ in Eq. \ref{eq:Q}, respectively \cite{li2020fast}.

\textbf{Matching step.} The feature keypoint matching problem can be described as a Gaussian Mixture Model (GMM), which is a classic and prevalent probability method in point matching and registration applications \cite{852377, 5432191, 5674050}. Since the two sets of points to be registered always have similar shape structures and conform to the same probability density model, GMM can be used to mine the structural relationship between the two point sets and ensure that the shapes represented by the registered point sets have correct corresponding structures \cite{YANG2015156}. For example, in \cite{5432191, 7185406, 9879560}, GMM is utilized to assign the point correspondence by estimating the parameters of the mixture model via a maximum likelihood and expectation maximization (EM) algorithms. Similarly, in this paper, we consider the edge contour points in Y as the centroids of a GMM and X as the data points generated by the GMM. Clothes shape matching can be describe as fitting the contour points X  to the points Y by maximizing a likelihood function. Specifically, $Z = \{ z_j \}, 1 \leq j \leq n$ represents the assignments of $\mathbf{y}_j$ to a Gaussian distribution with zero mean and $\sigma^{2}$ variance. The likelihood for $\mathbf{y}_j$ with a correspondence $\mathbf{x}_i$ is $P(\mathbf{y}_j | z_j = i) = N(\mathbf{y}_j | f(\mathbf{x}_i), \sigma^2)$.
In the matching step, the correspondence probability $p_{ij}$ for $\mathbf{y}_j$ and $\mathbf{x}_i$ can be approximated by \cite{li2020fast}
\begin{equation}\label{eq:q3}
	p_{ij}=\frac{\pi_{i} e^{-\frac{\| \mathbf{y}_j-f(\mathbf{x}_i) \|^2}{2\sigma^2}}}{ \sum_{i=1}^{m} \pi_{i} e^{-\frac{\| \mathbf{y}_j-f(\mathbf{x}_i) \|^2}{2\sigma^2}} + \frac{\gamma (2\pi \sigma^2)^{D/2}}{(1-\gamma)a} },
\end{equation}
where $\sum_{i=1}^m \pi_i=1$ are mixture weights, and $\sigma^2$ is the variance of the Gaussian distribution.

In order to obtain the edge points of the clothes, edge scanning is performed on the clothes mask $C_m$ and the clothes area mask $G_C$ respectively. It is worth noting that the density of the point cloud is not required to be as high as possible. High density point cloud data not only requires longer alignment time than low density point cloud data, but also is susceptible to noise during the registration process. In order to reduce the point cloud density, we sample points with large curvature changes.

\textbf{Warping step.} In the warping step, we utilize the Moving Least Squares (MLS) method \cite{schaefer2006image} to estimate the image deformation function $f$. Specifically, we use the collected cloth edge points as control points to drive the image deformation. Let $\mathbf{d}$ be the control points and $\mathbf{b}$ be the goal points, given a pixel point $v$ in an image, the goal of MLS is to find the best transformation $f_p$ that minimizes.

\begin{equation}
	\mathop{min}\limits_{f_p(\cdot)} E = \sum_i \mathbf{w}_i\vert f_p(\mathbf{d}_i)-\textbf{b}_i\vert^2,
\end{equation}
where $\mathbf{d}_i$ and $\mathbf{b}_i$ are row vectors, and the form of weights $\mathbf{w}_i$ is: 

\begin{equation}
	\mathbf{w}_i = {{1} \over {\vert \mathbf{d}_i - v\vert^{2\alpha}}},
\end{equation}
where $\alpha \le 1$ is the deformation tolerance.

In particular, the types of transformations used in MLS can be similarity transformations, affine transformations, or rigid transformations. In this paper, we estimate $f_p$ as similarity transformations.

\subsection{Image Synthesis Module (ISM)} \label{ISM}

In the virtual try-on task, it is not only necessary to preserve human features but also to adaptively generate missing body parts (e.g., missing arm regions due to changing from long sleeves to short sleeves). To solve this problem, we introduce the CGAN in ISM module. Specifically, ISM uses the outputs of the first two modules as inputs to synthesize the virtual try-on results by the CGAN network, and adaptively fills in missing areas. As shown in Fig. \ref{fig_3}, the input of ISM is ($W$, $S_r$, $I_a$). By combining the semantic information of $S_r$, CGAN integrates the distorted clothing image $W$ into the specified region of the clothing-agnostic human body image $I_a$ and adaptively generates the missing pose information. Finally, a virtual try-on result that preserves the detailed features is generated.

\begin{figure}[t!]
	\centering
	\begin{center}
		\includegraphics[width=0.95\linewidth]{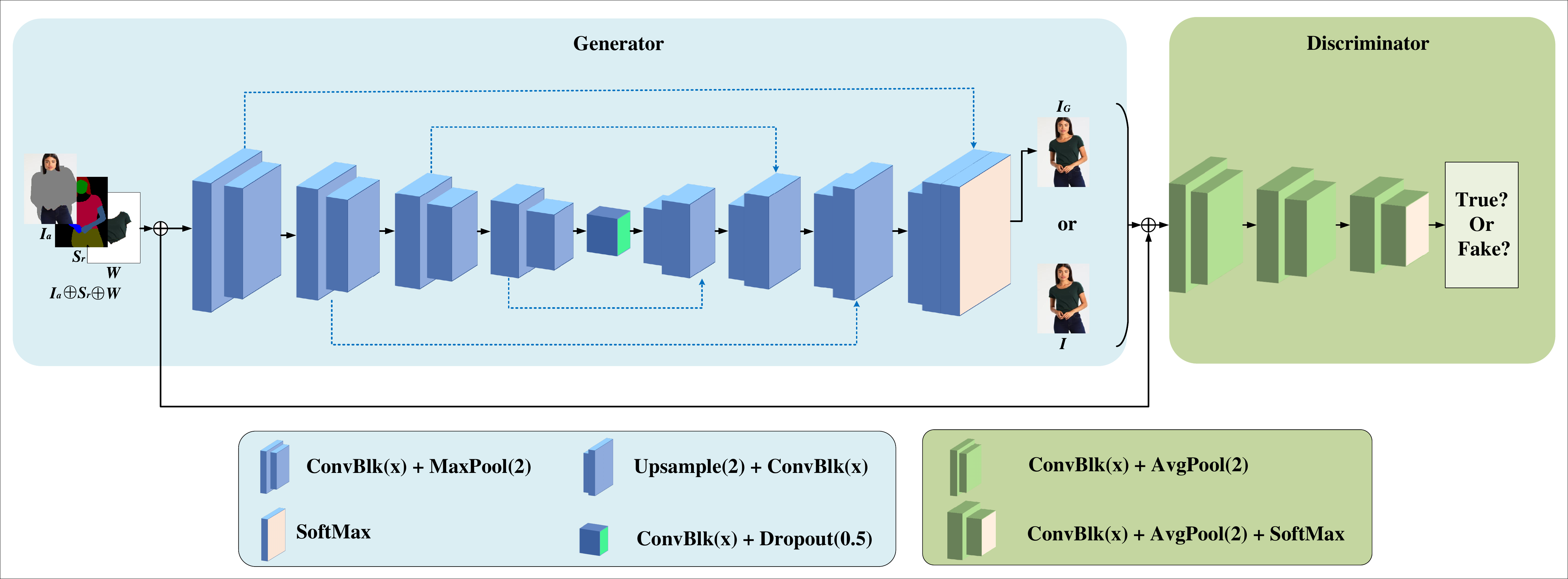}
	\end{center}
	\caption{Schematic diagram of the generator and discriminator of Image Synthesis Module.}
	\label{fig_5}
\end{figure}

\textbf{Model Architectures.} The CGAN used in VITON-DRR consists of a 9-layer U-net. As shown in Fig. \ref{fig_5}, the filter numbers in the down-sampling convolution layers are 64, 128, 256, 512, and 1024, while the filter numbers in the up-sampling convolution layers are 512, 256, 128, and 64, respectively.  Correspondingly, the discriminative network consists of three down-sampling convolution layers with filter numbers  64, 128, and 256, respectively.



\section{Experiments}
\subsection{Experiment Setup}
\subsubsection{Dataset}

\begin{figure}[]
	\centering
	\begin{center}
		\includegraphics[width=0.95\linewidth]{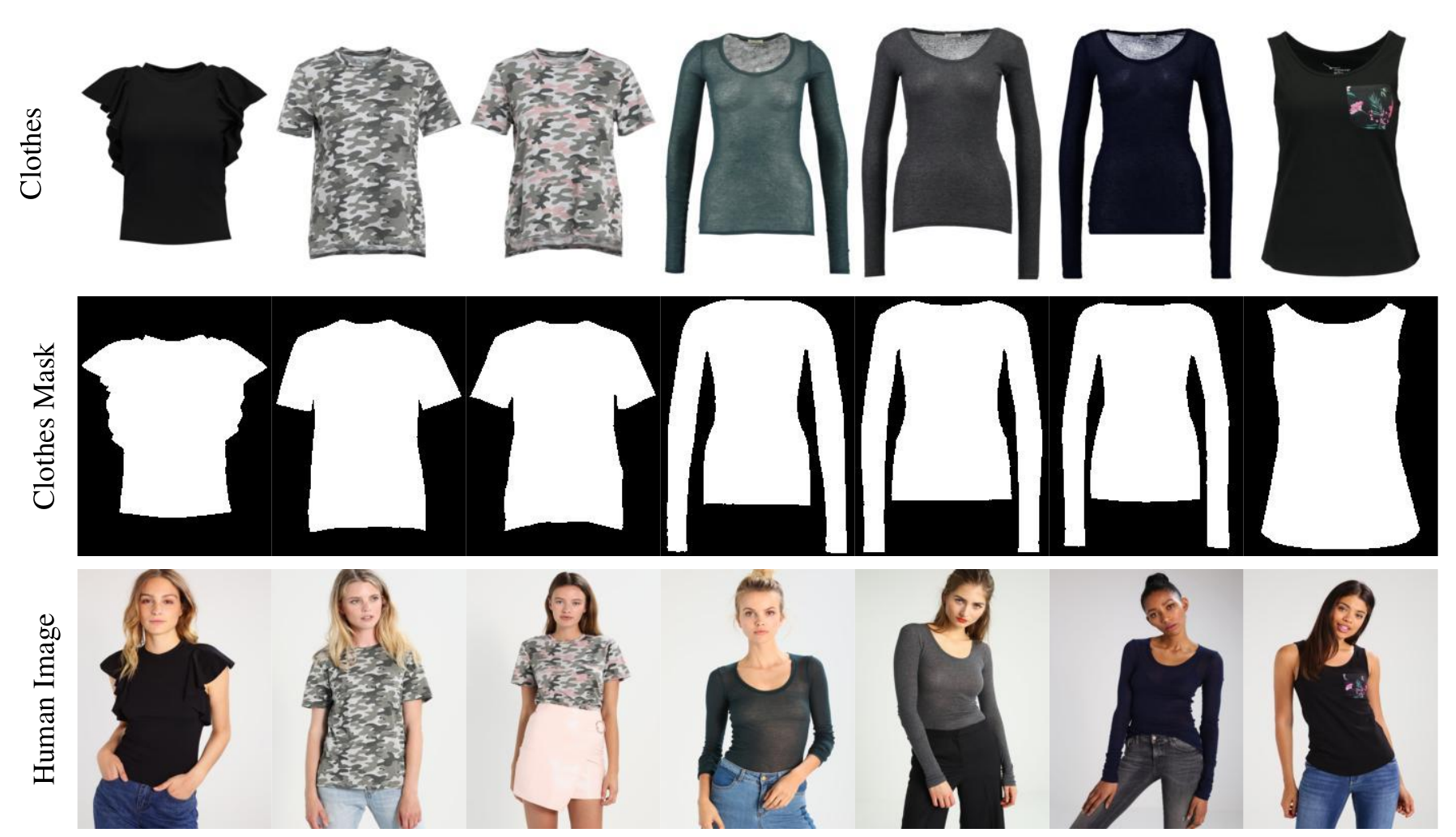}
	\end{center}
	\caption{Visualization of selected samples in the Zalando dataset.}
	\label{fig_6}
\end{figure}

For comparison with prior methods, we validate our method on the Zalando dataset, which is commonly used in many virtual try-on works \cite{han2018viton, wang2018toward, minar2020cp, yang2020towards, he2022style, bai2022single}. The Zalando dataset contains 16253 pairs of frontal photographs of females and their corresponding blouses. These photos are further divided into 14221 training pairs and 2032 test pairs. The image pairs in the test set are randomly scrambled to suit the actual situation.
Fig. \ref{fig_6} presents examples of the Zalando Dataset, which includes: (i) high-resolution clothes images (e.g., T-shirts and shirts) with transparent backgrounds,
(ii) associated semantic segmentation masks, and (iii) multi-pose human model images exhibiting diversity in body shape and posture.

\subsubsection{Training and Testing}
In our experiment, we train each module separately in stages. Each module is trained 50 epochs, except the deformation module, with loss weight $\lambda=1$ and batch size 8. The hyper-parameters of the Adam \cite{kingma2014adam} optimizer in our model are set to $\beta_1=0.5$, $ \beta_2=0.999$, and the learning rate is initialized to 0.0001. The resolution of the data inputs and the result outputs is $256 \times 192$. In DM, we use the Canny operator to sample the object edges of the image and the proposed non-rigid registration method to align the sampled salient edge points. In this way, we obtain the deformation parameters. In the test phase, we initialize the same experimental parameter settings. Most of our code is implemented using the deep learning toolkit Pytorch, and the point cloud registration code in DM consists of Matlab. We use an NVIDIA 3060 GPU in our experiment. Our code is publicly available.

\subsection{Qualitative Analysis}
To qualitatively evaluate our method, we visually compare the proposed method with state-of-the-art methods including CP-VITON \cite{wang2018toward}, CP-VITON+ \cite{minar2020cp}, ACGPN \cite{yang2020towards}, Flow-Style \cite{he2022style} and DAFlow \cite{bai2022single}.

\begin{figure}[t!]
	\centering
	\begin{center}
		\includegraphics[width=1\linewidth]{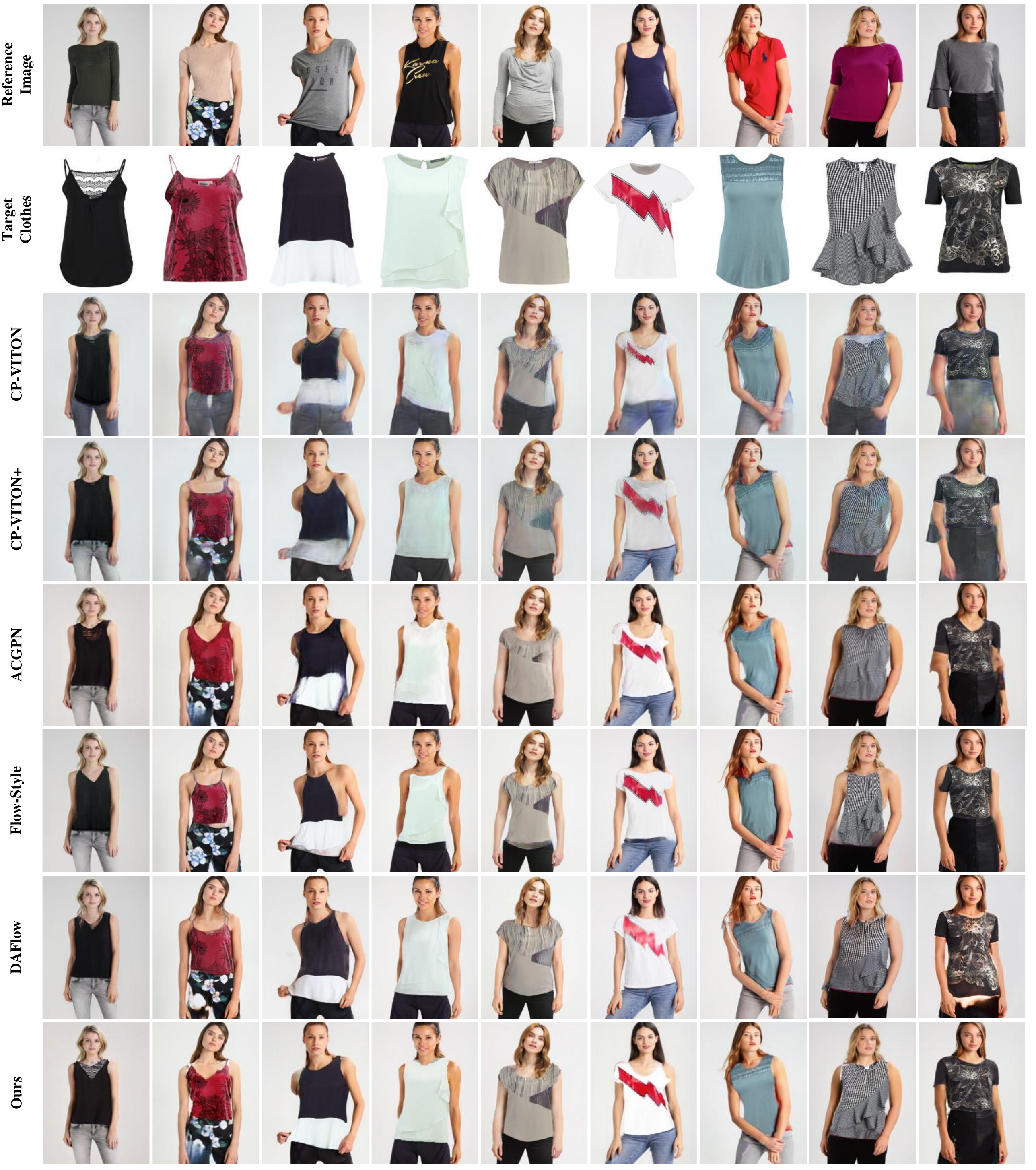}
	\end{center}
	\caption{Visual comparisons of CP-VITON, CP-VITON+, ACGPN, Flow-Style, DAFlow, and the proposed method. The virtual try-on results of different types of clothes (from sleeveless to short)  are presented. Our method retains more details because the images are precisely warped and the details are preserved for transfer to the appropriate human body area.}
	\label{fig_7}
\end{figure}

\textbf{Try-on Results.}
As shown in the first two columns of Fig. \ref{fig_7}, the baseline methods fail to preserve the stripe pattern after warping, especially around the shoulders. This poses a challenge in constructing correspondences between complex patterns such as stripes. In contrast, our VITON-DRR can accurately warp the stripe pattern to the target person and produce high-quality images even in complex poses. This is because our DM module first extracts dense keypoints and warps them precisely.

\begin{figure}[t!]
	\centering
	\begin{center}
		\includegraphics[width=1\linewidth]{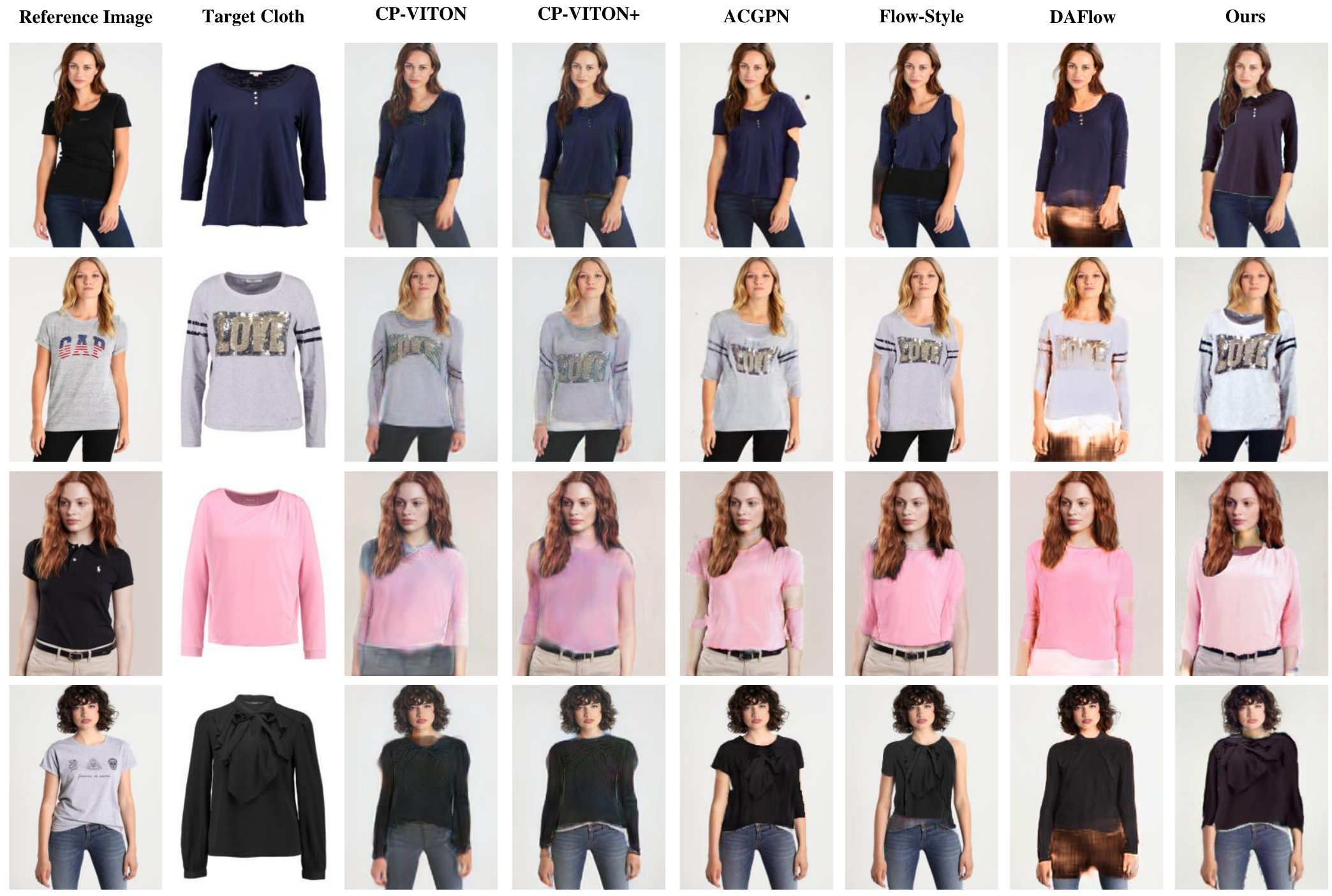}
	\end{center}
	\caption{Comparison between our method and existing methods in virtual try-on for long-sleeved garments. Compared to sleeveless and short-sleeved clothing, long-sleeved garments pose greater challenges due to their higher dynamic complexity. Although our method exhibits slightly inferior performance in detail preservation and stability compared to short-sleeved cases, it still effectively maintains the original garment’s shape, texture, and the target body retention regions.}
	\label{fig_8}
\end{figure}

The third and fourth columns show the results where the baseline methods fail to handle large deformations in different poses and result in blurred or incorrect try-on results. Specifically, CP-VITON and Flow-Style lead to misaligned clothing and blurred boundaries due to poor accuracy in matching the cloth to the body area. ACGPN achieves better results in cloth deformations but exhibits color artifacts at the boundary.

In addition, when complex poses appear in person images such as hand and finger occlusions, CP-VITON+ and DAFlow produce blurred results with unnatural distortions. Unlike them, our method preserves the characteristics of the target clothes and the body parts, benefiting from the robust non-rigid registration algorithm for occlusions in our DM. From the fifth to the ninth column, baseline models are also weak at preserving clothing details, such as the red pattern in the sixth column and the skirt wrinkles in the last column. Overall, our results are closer to the original garment images in terms of cloth integrity. This is because the non-rigid registration method used in our DM can accurately calculate the deformation parameters of images characterized by keypoints.

To demonstrate the generalization capability of our method for virtual try-on tasks, we present additional results for long-sleeved garments in  Fig. \ref{fig_8}. Compared to sleeveless and short-sleeved clothing, long-sleeved garments require higher accuracy in semantic segmentation, particularly in ensuring proper transfer of sleeve attributes to the target human images. Compared to other methods, our method performs better in preserving fine-grained garment details. Nevertheless, owing to the increased dynamic geometric complexity of long-sleeved garments and similar styles compared to short-sleeved counterparts, combined with the limited representation of such samples in current virtual try-on datasets, our method demonstrates comparatively reduced stability in handling certain garment styles under specific conditions.

\begin{figure}[h!]
	\centering
	\begin{center}
		\includegraphics[width=0.9 \linewidth]{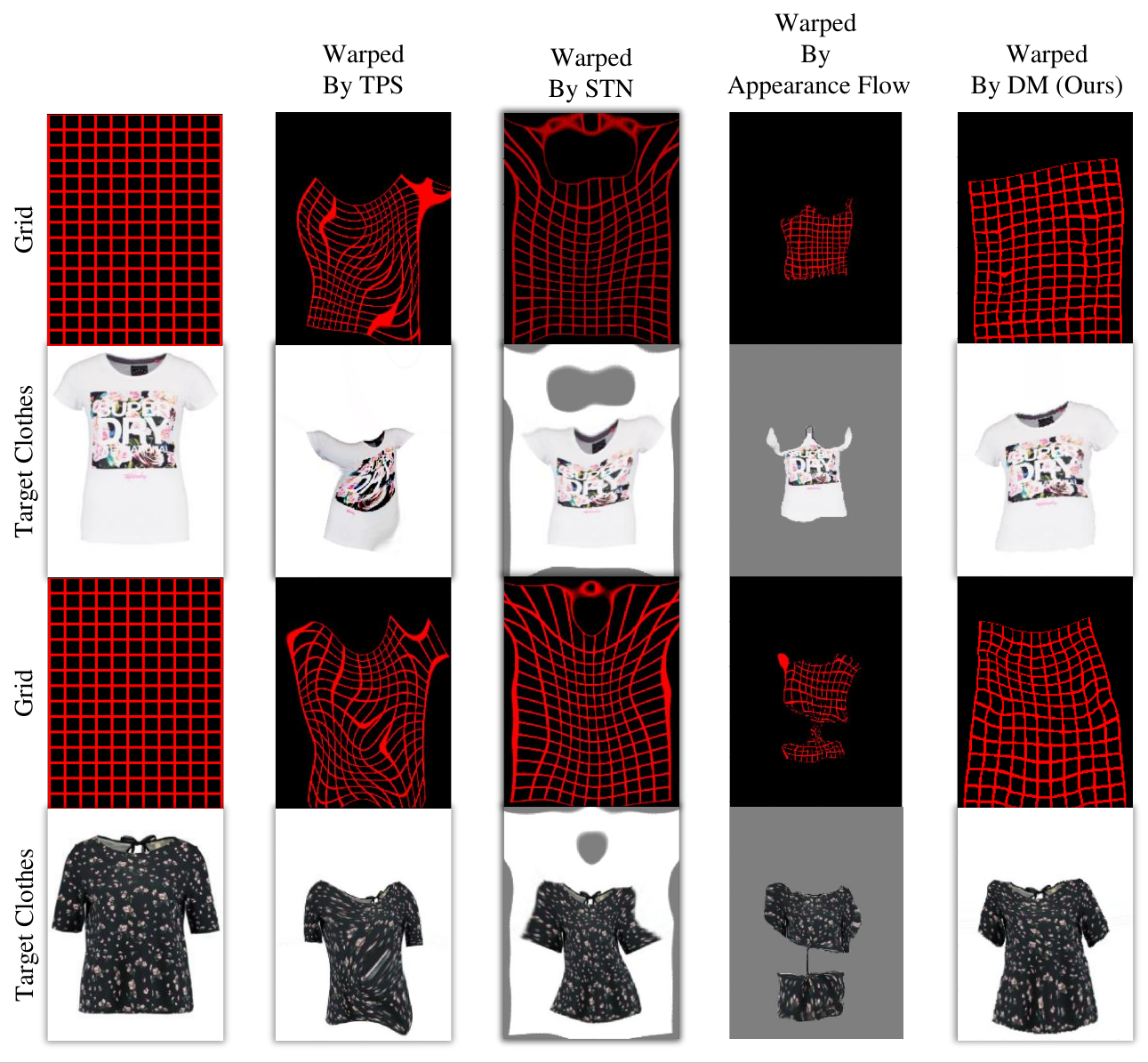}
	\end{center}
	\caption{Comparison of warping effects between TPS \cite{han2018viton}, STN \cite{yang2020towards} and Appearance Flow \cite{he2022style} and the proposed method. In our DM module, deformed images using the MLS method have a smoother effect without any local unreasonable deformation.}
	\label{fig_9}
\end{figure}

\begin{figure}[h!]
	\centering
	\begin{center}
		\includegraphics[width=0.65\linewidth]{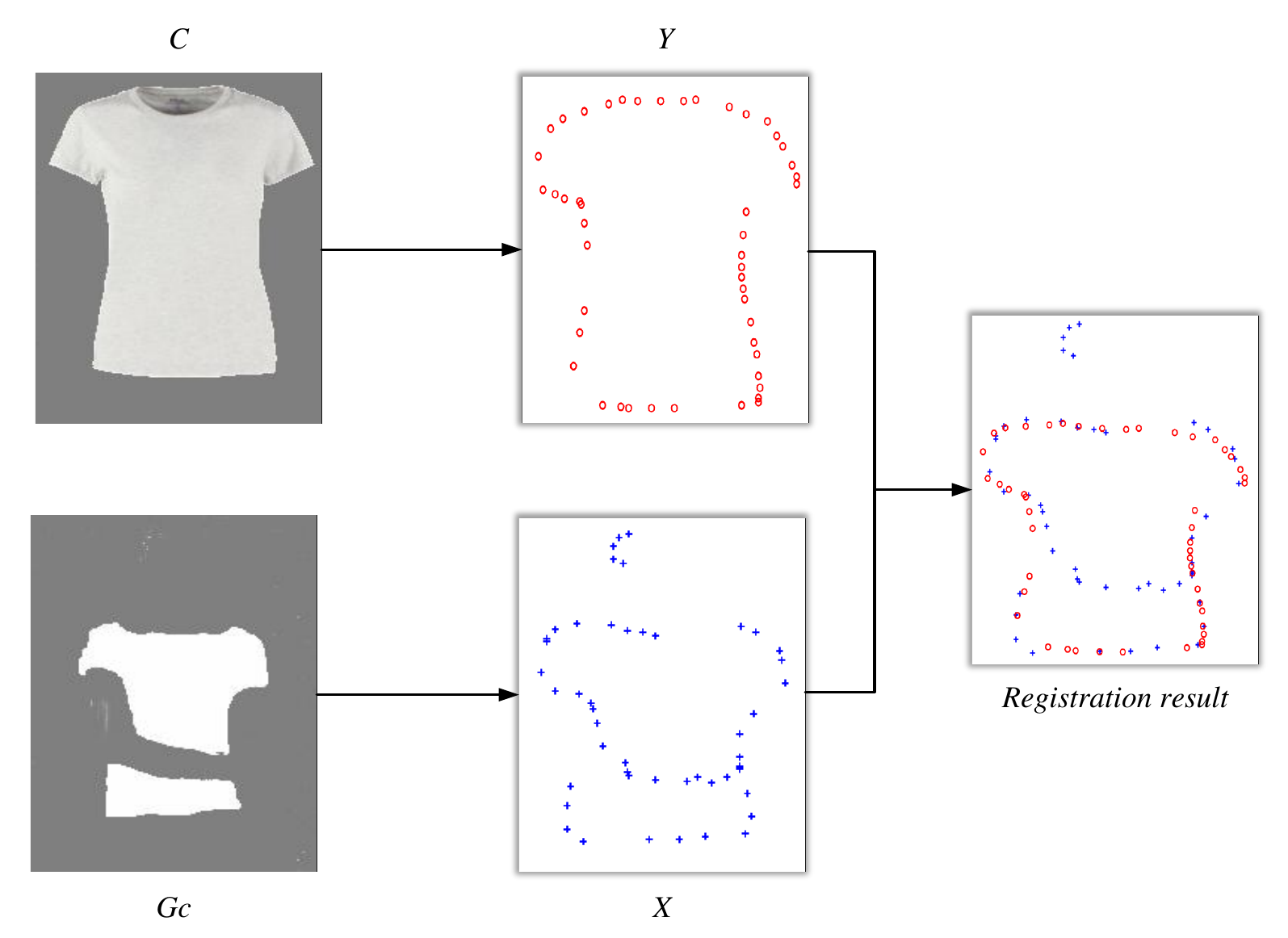}
	\end{center}
	\caption{{An example of the non-rigid point cloud registration with self-occlusions by our DM module.}}
	\label{fig_10}
\end{figure}

\textbf{Warping Results.} Under the condition of second-order difference constraints, the deformation effect of our DM method is better than that of other methods. As shown in Fig. \ref{fig_9}, the warping results of TPS may cause image distortion. Although the warping results obtained by STN improve the artifacts on the whole, the deformation of the garment is not reasonable. In the warping results of the appearance flow, the pattern on the white T-shirt appears to have a noticeable blur and loss of detail. Besides, the appearance flow method may lead to excessive deformation of the mesh around the waist of the dark patterned shirt, resulting in an unnatural shape. On the contrary, the warping obtained by our method do not produce artifacts and better ensure the integrity of the garment, which is more in line with the stretching of the garment in reality. In addition, in our warping stage, the proposed DM module can effectively address the self-occlusion issues of the clothing mask caused by the overlap of limbs and clothes, as shown in Fig. \ref{fig_10}. This is attributed to the robustness of our non-rigid registration algorithm.

\subsection{Quantitative Results}

To better illustrate the effectiveness of the proposed method, we conduct a quantitative evaluation of CP-VITON, CP-VITON+, ACGPN, Flow-Style, DAFlow and our proposed method using Inception Score (IS) \cite{salimans2016improved}, Structural Similarity (SSIM) \cite{wang2004image}, and Learned Perceptual Image Patch Similarity (LPIPS) \cite{zhang2018unreasonable}. The IS score is positively correlated with image quality, with higher scores indicating clear and diverse images. Similarly, higher   SSIM scores indicate higher similarity between two images. Differently, the LPIPS score is negatively correlated with image quality, with lower scores indicating closer similarity between two images. We expect that the characteristics of the virtual try-on results should be closer to the characteristics of the target clothing. Therefore, we measure the LPIPS between the virtual try-on results and the target clothes.

\begin{table}[th!]
	\caption{Quantitative comparison of CP-VITON, CP-VITON+, ACGPN, Flow-Style, DAFlow, and the proposed method using the three evaluation metrics. Higher scores are better for IS and SSIM, while the opposite is true for LPIPS. The red numbers represent the first rank, and the green numbers represent the second rank results. \label{tab:table1}}
	
	\scriptsize
	\begin{tabular*}{\hsize}{@{\extracolsep{\fill}}c c c c}
		\hline
		Method & IS $\uparrow$ & SSIM $\uparrow$ & LPIPS $\downarrow$\\
		\hline\hline
		CP-VITON & 1.242 & 0.741 & 0.587\\
		CP-VITON+ & 1.311 & 0.777 & 0.582\\
		ACGPN & 1.369 & 0.758 & 0.588\\
		Flow-Style & \textcolor{red}{\textbf{1.421}} & \textcolor{red}{\textbf{0.812}} & 0.590 \\
		DAFlow & 1.332 & 0.769 & \textcolor{green}{\textbf{0.577}}\\
		Ours & \textcolor{green}{\textbf{1.393}} &\textcolor{green}{\textbf{0.779}} &  \textcolor{red}{\textbf{0.575}}\\
		\hline
	\end{tabular*}
\end{table}

As shown in Tab. \ref{tab:table1}, CP-VITON provides baseline performance across all evaluation metrics. CP-VITON+ improves the composition mask by using CP-VITON's input garment mask and a specific loss function to obtain better results. Unfortunately, it has limitations when it comes to covering targets with different postures. DAFlow outperforms CP-VITON and CP-VITON+ on IS and LPIPS scores because it uses a deformable attention flow framework to preserve body features. However, it retains fewer details of the garment pattern. ACGPN has improved in retaining the details of body parts and the overall clarity of the garment. This is attributed to the redistribution of human semantic segmentation by the semantic generation module. 
However, ACGPN still suffers from some unnatural deformations and blurred garment patterns in preserving garment details. Compared with other methods, the global appearance Flow estimation model used by flow-style can guarantee more clothing details. However, the accuracy of clothing deformation is still not enough.

The proposed VITON-DRR method ranks second in the IS and SSIM evaluation metrics, and first in the LPIPS evaluation metric. Through efficient non-rigid registration, VITON-DRR makes the image deformation smoother and more accurate while preserving more garment details. Overall, the proposed VITON-DRR method provides more balanced fitting results in terms of deformation accuracy and garment detail retention.

\subsection{Computation Analysis}
As shown in Table 2, to enhance the model's ability to handle complex tasks and fine-grained details, the number of parameters in our model (47.61 M) has increased compared to other methods. However, while maintaining accuracy, our model's FLOPs (floating-point operations) remain competitive. For instance, the FLOP count of our method (13.75 G) is comparable to those of CP-VITON and CP-VITON+, while being significantly lower than ACGPN (49.27 G), FlowStyle (44.51 G), and DA-Flow (74.70 G).

In terms of computational efficiency, our model demonstrates superior performance compared to most existing methods, with an inference speed only slightly slower than FlowStyle. This evidence indicates that our approach maintains relatively low computational complexity while handling pose variations, thereby achieving an optimal balance between processing efficiency and output quality.

\begin{table}[]
	\caption{Comparison of computational efficiency across different methods.} \label{tab:table2}
	\scriptsize
	\begin{tabular*}{\hsize}{@{\extracolsep{\fill}}c c c c c c c c}
		\hline
		 & CP-VITON  & CP-VITON+ & ACGPN & FlowStyle & DA-Flow & Ours\\
		\hline\hline
		FLOPs (G) & 8.59 & 8.61 & 49.27 & 44.51 & 74.70 & 13.75\\
		Params (M) & 21.35 & 21.35 & 36.47 & 32.32 & 37.65 & 47.61\\
		Runtime (s) & 5.43 & 3.64 & 3.71 & 2.52 & 3.54 & 2.96\\
		\hline
	\end{tabular*}
\end{table}
\subsection{Ablation Study}

\begin{figure}[th!]
	\centering
	\begin{center}
		\includegraphics[width=0.8\linewidth]{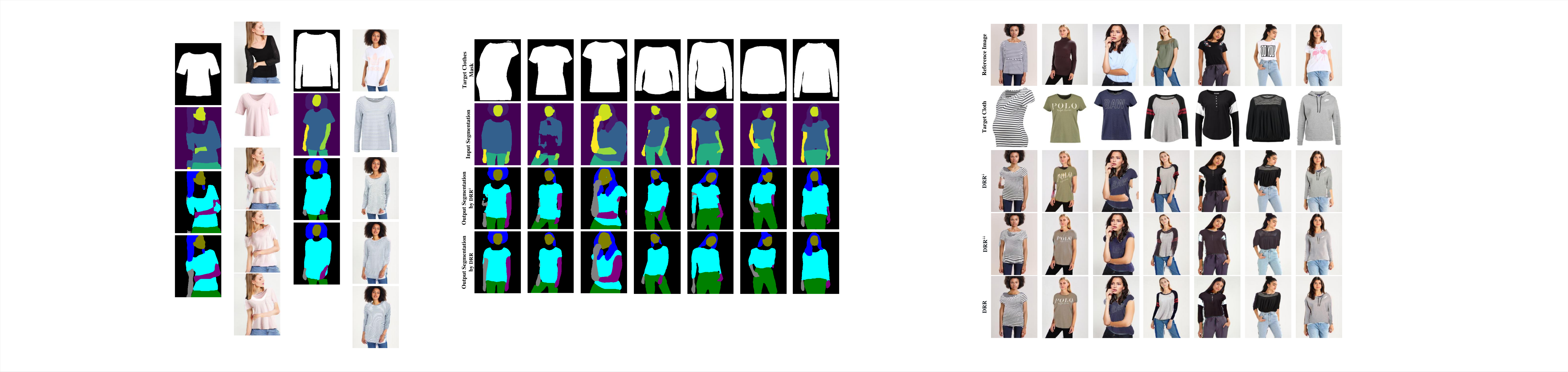}
	\end{center}
	\caption{Impact of the human semantic segmentation module in the ablation study.}
	\label{fig_11}
\end{figure}

To demonstrate the effectiveness of the proposed two modules in the overall method, we designed two control groups, DRR$\dagger$ and DRR$\dagger\dagger$, in the ablation experiments. Specifically, DRR$\dagger$ replaces our HSGM module with a GAN network, while DRR$\dagger\dagger$ replaces our DM module with the STN $+$ TPS algorithm.  ACGPN serves as the baseline model for this ablation study. 

Fig. \ref{fig_11} shows the semantic segmentation results of the proposed HSGM module on the human body under the given condition of a clothing mask. The results demonstrate that our method can effectively simulate the hidden or visible state of human limbs, regardless of long or short-sleeved clothing. Even when the artificial semantic segmentation of the input is incomplete, our method can still efficiently complete the missing parts. In contrast, the baseline model is deficient in accurately generating human semantic segmentation maps.

\begin{figure}[h!]
	\centering
	\begin{center}
		\includegraphics[width=0.8\linewidth]{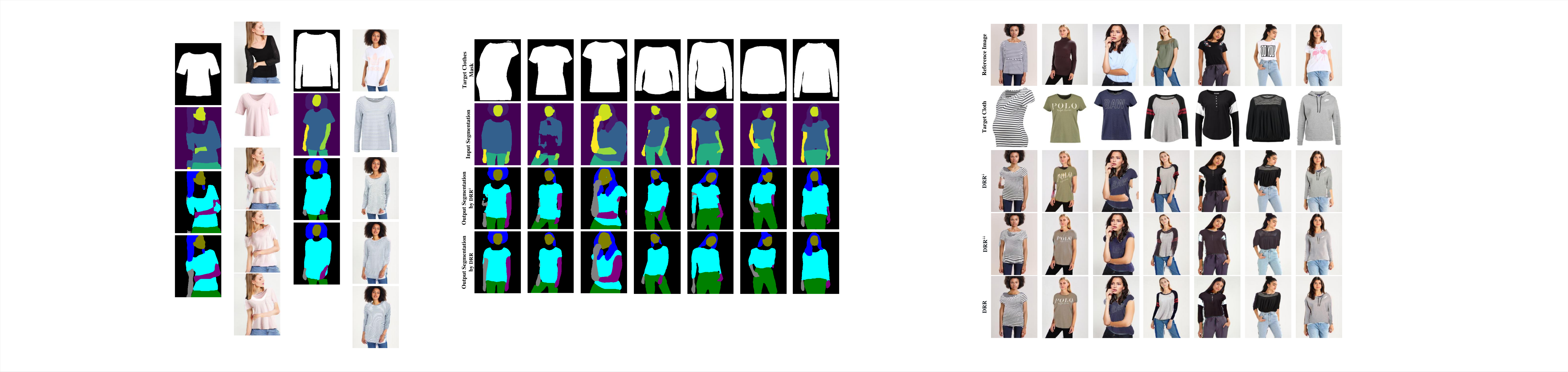}
	\end{center}
	\caption{Visual comparison of virtual try-on results with and without the HSGM and the DM.}
	\label{fig_12}
\end{figure}

The quality of virtual try-on results depends critically on both the accuracy of semantic segmentation maps and the effectiveness of image warping. Fig. \ref{fig_12} presents comparative results with and without the DM module and HSGM module. Without employing the HSGM module, DRR$\dagger$ demonstrates inferior performance. With more accurate semantic segmentation maps, DRR$\dagger\dagger$ achieves relatively better results. However, the absence of the DM module leads to loss of fine details in garment warping in DRR$\dagger\dagger$. In contrast, DRR not only generates precise semantic segmentation maps but also preserves detailed patterns in warped garments, thereby ensuring accurate clothing image generation.

Quantitative comparisons of the ablation study are presented in  Tab. \ref{tab:table3}. Our complete method achieves superior performance across all control groups.

\begin{table}[]
	\caption{Quantitative comparisons between our method and the control groups.} \label{tab:table3}
	
	\scriptsize
	\begin{tabular*}{\hsize}{@{\extracolsep{\fill}}c c c c}
		\hline
		Method & IS $\uparrow$ & SSIM $\uparrow$ & LPIPS $\downarrow$\\
		\hline\hline
		DRR$\dagger\dagger$ & 1.352 & 0.728 & 0.600\\
		DRR$\dagger$ & 1.357 & 0.754 & 0.595\\
		DRR & \textcolor{red}{\textbf{1.393}} & \textcolor{red}{\textbf{0.779}} & \textcolor{red}{\textbf{0.575}}\\

		\hline
	\end{tabular*}
\end{table}

\section{Conclusions and Discussion}
In this paper, we have proposed a detail retention virtual try-on model (VITON-DRR), which focuses on preserving clothing details. Our framework consists of three modules, i.e., the Human Semantics Generation Module (HSGM), the Deformation Module (DM), and the Image Synthesis Module (ISM). By introducing an innovative non-rigid registration-based approach to jointly learn target garment matching and warping, our model achieves smoother and more accurate image deformations while retaining finer garment details, thereby producing high-quality fitting results. We evaluate VITON-DRR quantitatively and qualitatively on the Zalando dataset. Experimental results demonstrate that our method outperforms state-of-the-art approaches in garment detail preservation.

However, garments with long sleeves and other complex styles exhibit higher dynamic complexity, posing additional challenges. Furthermore, existing virtual try-on datasets lack sufficient samples for certain garment styles, leading to suboptimal parameter learning and reduced performance for these categories. In future work, we plan to integrate cross-modal models (e.g., CLIP) to address these limitations and enhance generalization.


\section{Acknowledgments}
This work was partially supported by the Chinese National Natural Science Foundation of China under Grant No. 61971339, the Natural Science Foundation of Shaanxi Province, China under Grant 2022JM-348 and 2022JM-394, and in part by the Doctoral Startup Foundation of Xian Polytechnic University under Grant BS1726.


\bibliographystyle{elsarticle-harv}
\bibliography{reference.bib}

\begin{thebibliography}{58}
\expandafter\ifx\csname natexlab\endcsname\relax\def\natexlab#1{#1}\fi
\providecommand{\url}[1]{\texttt{#1}}
\providecommand{\href}[2]{#2}
\providecommand{\path}[1]{#1}
\providecommand{\DOIprefix}{doi:}
\providecommand{\ArXivprefix}{arXiv:}
\providecommand{\URLprefix}{URL: }
\providecommand{\Pubmedprefix}{pmid:}
\providecommand{\doi}[1]{\href{http://dx.doi.org/#1}{\path{#1}}}
\providecommand{\Pubmed}[1]{\href{pmid:#1}{\path{#1}}}
\providecommand{\bibinfo}[2]{#2}
\ifx\xfnm\relax \def\xfnm[#1]{\unskip,\space#1}\fi
\bibitem[{Adhikari et~al.(2023)Adhikari, Bhusal, Ghimire and
  Shrestha}]{adhikari2023vton}
\bibinfo{author}{Adhikari, S.}, \bibinfo{author}{Bhusal, B.},
  \bibinfo{author}{Ghimire, P.}, \bibinfo{author}{Shrestha, A.},
  \bibinfo{year}{2023}.
\newblock \bibinfo{title}{Vton-it: Virtual try-on using image translation}.
\newblock \bibinfo{journal}{arXiv preprint arXiv:2310.04558} .
\bibitem[{Bai et~al.(2022)Bai, Zhou, Li, Zhou and Yang}]{bai2022single}
\bibinfo{author}{Bai, S.}, \bibinfo{author}{Zhou, H.}, \bibinfo{author}{Li,
  Z.}, \bibinfo{author}{Zhou, C.}, \bibinfo{author}{Yang, H.},
  \bibinfo{year}{2022}.
\newblock \bibinfo{title}{Single stage virtual try-on via deformable attention
  flows}, in: \bibinfo{booktitle}{Proc. of the European Conference on Computer
  Vision (ECCV)}, \bibinfo{organization}{Springer}. pp.
  \bibinfo{pages}{409--425}.
\bibitem[{Bhatnagar et~al.(2019)Bhatnagar, Tiwari, Theobalt and
  Pons-Mo}]{bhatnagar2019multi}
\bibinfo{author}{Bhatnagar, B.L.}, \bibinfo{author}{Tiwari, G.},
  \bibinfo{author}{Theobalt, C.}, \bibinfo{author}{Pons-Mo},
  \bibinfo{year}{2019}.
\newblock \bibinfo{title}{Multi-garment net: Learning to dress 3d people from
  images}, in: \bibinfo{booktitle}{{IEEE} International Conference on Computer
  Vision ({ICCV})}, \bibinfo{organization}{{IEEE}}. pp.
  \bibinfo{pages}{5420--5430}.
\bibitem[{Bookstein(1989)}]{bookstein1989principal}
\bibinfo{author}{Bookstein, F.L.}, \bibinfo{year}{1989}.
\newblock \bibinfo{title}{Principal warps: Thin-plate splines and the
  decomposition of deformations}.
\newblock \bibinfo{journal}{In IEEE Trans. on Pattern Analysis and Machine
  Intelligence (PAMI)} \bibinfo{volume}{11}, \bibinfo{pages}{567--585}.
\bibitem[{Bosquet et~al.(2023)Bosquet, Cores, Seidenari, Brea, Mucientes and
  Del~Bimbo}]{bosquet2023full}
\bibinfo{author}{Bosquet, B.}, \bibinfo{author}{Cores, D.},
  \bibinfo{author}{Seidenari, L.}, \bibinfo{author}{Brea, V.M.},
  \bibinfo{author}{Mucientes, M.}, \bibinfo{author}{Del~Bimbo, A.},
  \bibinfo{year}{2023}.
\newblock \bibinfo{title}{A full data augmentation pipeline for small object
  detection based on generative adversarial networks}.
\newblock \bibinfo{journal}{Pattern Recognition} \bibinfo{volume}{133},
  \bibinfo{pages}{108998}.
\bibitem[{Bouchard et~al.(2023)Bouchard, Wiesner, Desch{\^e}nes, Bilodeau,
  Turcotte, Gagn{\'e} and Lavoie-Cardinal}]{bouchard2023resolution}
\bibinfo{author}{Bouchard, C.}, \bibinfo{author}{Wiesner, T.},
  \bibinfo{author}{Desch{\^e}nes, A.}, \bibinfo{author}{Bilodeau, A.},
  \bibinfo{author}{Turcotte, B.}, \bibinfo{author}{Gagn{\'e}, C.},
  \bibinfo{author}{Lavoie-Cardinal, F.}, \bibinfo{year}{2023}.
\newblock \bibinfo{title}{Resolution enhancement with a task-assisted gan to
  guide optical nanoscopy image analysis and acquisition}.
\newblock \bibinfo{journal}{Nature Machine Intelligence} \bibinfo{volume}{5},
  \bibinfo{pages}{830--844}.
\bibitem[{Brouet et~al.(2012)Brouet, Sheffer, Boissieux and
  Cani}]{brouet2012design}
\bibinfo{author}{Brouet, R.}, \bibinfo{author}{Sheffer, A.},
  \bibinfo{author}{Boissieux, L.}, \bibinfo{author}{Cani, M.P.},
  \bibinfo{year}{2012}.
\newblock \bibinfo{title}{Design preserving garment transfer}.
\newblock \bibinfo{journal}{ACM Trans. on Graphics} \bibinfo{volume}{31},
  \bibinfo{pages}{1--11}.
\bibitem[{Chen et~al.(2024a)Chen, Huang, Huang, Ge and
  Shao}]{chen2024gaussianvton}
\bibinfo{author}{Chen, H.}, \bibinfo{author}{Huang, Y.},
  \bibinfo{author}{Huang, H.}, \bibinfo{author}{Ge, X.}, \bibinfo{author}{Shao,
  D.}, \bibinfo{year}{2024}a.
\newblock \bibinfo{title}{Gaussianvton: 3d human virtual try-on via multi-stage
  gaussian splatting editing with image prompting}.
\newblock \bibinfo{journal}{arXiv preprint arXiv:2405.07472} .
\bibitem[{Chen et~al.(2024b)Chen, Zhang, Ma, Yang and Zhang}]{chen2024cs}
\bibinfo{author}{Chen, J.}, \bibinfo{author}{Zhang, X.}, \bibinfo{author}{Ma,
  L.}, \bibinfo{author}{Yang, B.}, \bibinfo{author}{Zhang, K.},
  \bibinfo{year}{2024}b.
\newblock \bibinfo{title}{Cs-viton: a realistic virtual try-on network based on
  clothing region alignment and spm}.
\newblock \bibinfo{journal}{The Visual Computer} , \bibinfo{pages}{1--15}.
\bibitem[{Chen and Ren(2024)}]{chen2024integrating}
\bibinfo{author}{Chen, T.}, \bibinfo{author}{Ren, J.}, \bibinfo{year}{2024}.
\newblock \bibinfo{title}{Integrating gan and texture synthesis for enhanced
  road damage detection}.
\newblock \bibinfo{journal}{IEEE Trans. on Intelligent Transportation Systems
  (T-ITS)} \bibinfo{volume}{25}, \bibinfo{pages}{12361--12371}.
\bibitem[{Cherian et~al.(2024)Cherian, Vaidhehi, Arshey, Briskilal and
  Simpson}]{cherian2024generative}
\bibinfo{author}{Cherian, A.K.}, \bibinfo{author}{Vaidhehi, M.},
  \bibinfo{author}{Arshey, M.}, \bibinfo{author}{Briskilal, J.},
  \bibinfo{author}{Simpson, S.V.}, \bibinfo{year}{2024}.
\newblock \bibinfo{title}{Generative adversarial networks with stochastic
  gradient descent with momentum algorithm for video-based facial expression}.
\newblock \bibinfo{journal}{International Journal of Information Technology}
  \bibinfo{volume}{16}, \bibinfo{pages}{3703--3722}.
\bibitem[{Chopra et~al.(2021)Chopra, Jain, Hemani and
  Krishnamurthy}]{chopra2021zflow}
\bibinfo{author}{Chopra, A.}, \bibinfo{author}{Jain, R.},
  \bibinfo{author}{Hemani, M.}, \bibinfo{author}{Krishnamurthy, B.},
  \bibinfo{year}{2021}.
\newblock \bibinfo{title}{Zflow: Gated appearance flow-based virtual try-on
  with 3d priors}, in: \bibinfo{booktitle}{{IEEE} International Conference on
  Computer Vision ({ICCV})}, \bibinfo{organization}{{IEEE}}. pp.
  \bibinfo{pages}{5433--5442}.
\bibitem[{Chui and Rangarajan(2000)}]{852377}
\bibinfo{author}{Chui, H.}, \bibinfo{author}{Rangarajan, A.},
  \bibinfo{year}{2000}.
\newblock \bibinfo{title}{A feature registration framework using mixture
  models}, in: \bibinfo{booktitle}{Proceedings IEEE Workshop on Mathematical
  Methods in Biomedical Image Analysis (MMBIA)}, pp. \bibinfo{pages}{190--197}.
\bibitem[{Chui and Rangarajan(2003)}]{chui2003new}
\bibinfo{author}{Chui, H.}, \bibinfo{author}{Rangarajan, A.},
  \bibinfo{year}{2003}.
\newblock \bibinfo{title}{A new point matching algorithm for non-rigid
  registration}.
\newblock \bibinfo{journal}{Computer Vision and Image Understanding}
  \bibinfo{volume}{89}, \bibinfo{pages}{114--141}.
\bibitem[{Delavari et~al.(2019)Delavari, Foruzan and
  Chen}]{delavari2019accurate}
\bibinfo{author}{Delavari, M.}, \bibinfo{author}{Foruzan, A.H.},
  \bibinfo{author}{Chen, Y.W.}, \bibinfo{year}{2019}.
\newblock \bibinfo{title}{Accurate point correspondences using a modified
  coherent point drift algorithm}.
\newblock \bibinfo{journal}{Biomedical Signal Processing and Control}
  \bibinfo{volume}{52}, \bibinfo{pages}{429--444}.
\bibitem[{Dong et~al.(2019)Dong, Liang, Shen, Wang, Lai, Zhu, Hu and
  Yin}]{dong2019towards}
\bibinfo{author}{Dong, H.}, \bibinfo{author}{Liang, X.}, \bibinfo{author}{Shen,
  X.}, \bibinfo{author}{Wang, B.}, \bibinfo{author}{Lai, H.},
  \bibinfo{author}{Zhu, J.}, \bibinfo{author}{Hu, Z.}, \bibinfo{author}{Yin,
  J.}, \bibinfo{year}{2019}.
\newblock \bibinfo{title}{Towards multi-pose guided virtual try-on network},
  in: \bibinfo{booktitle}{{IEEE} International Conference on Computer Vision
  ({ICCV})}, \bibinfo{organization}{{IEEE}}. pp. \bibinfo{pages}{9026--9035}.
\bibitem[{Du et~al.(2013)Du, Geng, Li and He}]{du20133d}
\bibinfo{author}{Du, H.}, \bibinfo{author}{Geng, G.}, \bibinfo{author}{Li, K.},
  \bibinfo{author}{He, Y.}, \bibinfo{year}{2013}.
\newblock \bibinfo{title}{3d skull registration based on registration points
  automatic correspondence}, in: \bibinfo{booktitle}{2013 International
  Conference on Virtual Reality and Visualization (ICVRV)},
  \bibinfo{organization}{IEEE}. pp. \bibinfo{pages}{293--296}.
\bibitem[{Fan et~al.(2022)Fan, Ma, Tian, Mei and Lin}]{9879560}
\bibinfo{author}{Fan, A.}, \bibinfo{author}{Ma, J.}, \bibinfo{author}{Tian,
  X.}, \bibinfo{author}{Mei, X.}, \bibinfo{author}{Lin, W.},
  \bibinfo{year}{2022}.
\newblock \bibinfo{title}{Coherent point drift revisited for non-rigid shape
  matching and registration}, in: \bibinfo{booktitle}{{IEEE} Conf. on Computer
  Vision and Pattern Recognition (CVPR)}, \bibinfo{organization}{{IEEE}}. pp.
  \bibinfo{pages}{1414--1424}.
\bibitem[{Ge et~al.(2021)Ge, Song, Zhang, Ge, Liu and Luo}]{ge2021parser}
\bibinfo{author}{Ge, Y.}, \bibinfo{author}{Song, Y.}, \bibinfo{author}{Zhang,
  R.}, \bibinfo{author}{Ge, C.}, \bibinfo{author}{Liu, W.},
  \bibinfo{author}{Luo, P.}, \bibinfo{year}{2021}.
\newblock \bibinfo{title}{Parser-free virtual try-on via distilling appearance
  flows}, in: \bibinfo{booktitle}{{IEEE} Conf. on Computer Vision and Pattern
  Recognition (CVPR)}, \bibinfo{organization}{{IEEE}}. pp.
  \bibinfo{pages}{8485--8493}.
\bibitem[{G{\"u}ler et~al.(2018)G{\"u}ler, Neverova and
  Kokkinos}]{guler2018densepose}
\bibinfo{author}{G{\"u}ler, R.A.}, \bibinfo{author}{Neverova, N.},
  \bibinfo{author}{Kokkinos, I.}, \bibinfo{year}{2018}.
\newblock \bibinfo{title}{Densepose: Dense human pose estimation in the wild},
  in: \bibinfo{booktitle}{{IEEE} Conf. on Computer Vision and Pattern
  Recognition (CVPR)}, \bibinfo{organization}{{IEEE}}. pp.
  \bibinfo{pages}{7297--7306}.
\bibitem[{Han et~al.(2019)Han, Hu, Huang and Scott}]{han2019clothflow}
\bibinfo{author}{Han, X.}, \bibinfo{author}{Hu, X.}, \bibinfo{author}{Huang,
  W.}, \bibinfo{author}{Scott, M.R.}, \bibinfo{year}{2019}.
\newblock \bibinfo{title}{Clothflow: A flow-based model for clothed person
  generation}, in: \bibinfo{booktitle}{{IEEE} International Conference on
  Computer Vision ({ICCV})}, \bibinfo{organization}{{IEEE}}. pp.
  \bibinfo{pages}{10471--10480}.
\bibitem[{Han et~al.(2018)Han, Wu, Wu, Yu and Da}]{han2018viton}
\bibinfo{author}{Han, X.}, \bibinfo{author}{Wu, Z.}, \bibinfo{author}{Wu, Z.},
  \bibinfo{author}{Yu, R.}, \bibinfo{author}{Da}, \bibinfo{year}{2018}.
\newblock \bibinfo{title}{Viton: An image-based virtual try-on network}, in:
  \bibinfo{booktitle}{{IEEE} Conf. on Computer Vision and Pattern Recognition
  (CVPR)}, \bibinfo{organization}{{IEEE}}. pp. \bibinfo{pages}{7543--7552}.
\bibitem[{He et~al.(2022)He, Song and Xiang}]{he2022style}
\bibinfo{author}{He, S.}, \bibinfo{author}{Song, Y.Z.}, \bibinfo{author}{Xiang,
  T.}, \bibinfo{year}{2022}.
\newblock \bibinfo{title}{Style-based global appearance flow for virtual
  try-on}, in: \bibinfo{booktitle}{{IEEE} Conf. on Computer Vision and Pattern
  Recognition (CVPR)}, \bibinfo{organization}{{IEEE}}. pp.
  \bibinfo{pages}{3470--3479}.
\bibitem[{Huang et~al.(2022)Huang, Li, Xie, Kampffmeyer, Liang
  et~al.}]{huang2022towards}
\bibinfo{author}{Huang, Z.}, \bibinfo{author}{Li, H.}, \bibinfo{author}{Xie,
  Z.}, \bibinfo{author}{Kampffmeyer, M.}, \bibinfo{author}{Liang, X.}, et~al.,
  \bibinfo{year}{2022}.
\newblock \bibinfo{title}{Towards hard-pose virtual try-on via 3d-aware global
  correspondence learning}, in: \bibinfo{booktitle}{Advances in Neural
  Information Processing Systems (NIPS)}, \bibinfo{organization}{Springer}. pp.
  \bibinfo{pages}{32736--32748}.
\bibitem[{Ishikawa and Ikenaga(2022)}]{ishikawa2022image}
\bibinfo{author}{Ishikawa, S.}, \bibinfo{author}{Ikenaga, T.},
  \bibinfo{year}{2022}.
\newblock \bibinfo{title}{Image-based virtual try-on system with clothing
  extraction module that adapts to any posture}.
\newblock \bibinfo{journal}{Computers \& Graphics} \bibinfo{volume}{106},
  \bibinfo{pages}{161--173}.
\bibitem[{Jetchev and Bergmann(2017)}]{jetchev2017conditional}
\bibinfo{author}{Jetchev, N.}, \bibinfo{author}{Bergmann, U.},
  \bibinfo{year}{2017}.
\newblock \bibinfo{title}{The conditional analogy gan: Swapping fashion
  articles on people images}, in: \bibinfo{booktitle}{{IEEE} International
  Conference on Computer Vision Workshops (ICCVW)},
  \bibinfo{organization}{{IEEE}}. pp. \bibinfo{pages}{2287--2292}.
\bibitem[{Jian and Vemuri(2011)}]{5674050}
\bibinfo{author}{Jian, B.}, \bibinfo{author}{Vemuri, B.C.},
  \bibinfo{year}{2011}.
\newblock \bibinfo{title}{Robust point set registration using gaussian mixture
  models}.
\newblock \bibinfo{journal}{IEEE Transactions on Pattern Analysis and Machine
  Intelligence} \bibinfo{volume}{33}, \bibinfo{pages}{1633--1645}.
\bibitem[{Kingma and Ba(2014)}]{kingma2014adam}
\bibinfo{author}{Kingma, D.P.}, \bibinfo{author}{Ba, J.}, \bibinfo{year}{2014}.
\newblock \bibinfo{title}{Adam: A method for stochastic optimization}.
\newblock \bibinfo{journal}{arXiv preprint arXiv:1412.6980} .
\bibitem[{Lee et~al.(2022)Lee, Gu, Park, Choi and Choo}]{lee2022high}
\bibinfo{author}{Lee, S.}, \bibinfo{author}{Gu, G.}, \bibinfo{author}{Park,
  S.}, \bibinfo{author}{Choi, S.}, \bibinfo{author}{Choo, J.},
  \bibinfo{year}{2022}.
\newblock \bibinfo{title}{High-resolution virtual try-on with misalignment and
  occlusion-handled conditions}, in: \bibinfo{booktitle}{Proc. of the European
  Conference on Computer Vision (ECCV)}, \bibinfo{organization}{Springer}. pp.
  \bibinfo{pages}{204--219}.
\bibitem[{Li et~al.(2020)Li, Xu, Xin, Zhang and Jing}]{li2020fast}
\bibinfo{author}{Li, M.}, \bibinfo{author}{Xu, R.Y.}, \bibinfo{author}{Xin,
  J.}, \bibinfo{author}{Zhang, K.}, \bibinfo{author}{Jing, J.},
  \bibinfo{year}{2020}.
\newblock \bibinfo{title}{Fast non-rigid points registration with cluster
  correspondences projection}.
\newblock \bibinfo{journal}{Signal Processing} \bibinfo{volume}{170},
  \bibinfo{pages}{107425}.
\bibitem[{Lin and Caesar(2024)}]{lin2024icp}
\bibinfo{author}{Lin, Y.}, \bibinfo{author}{Caesar, H.}, \bibinfo{year}{2024}.
\newblock \bibinfo{title}{Icp-flow: Lidar scene flow estimation with icp}, in:
  \bibinfo{booktitle}{{IEEE} Conf. on Computer Vision and Pattern Recognition
  (CVPR)}, \bibinfo{organization}{{IEEE}}. pp. \bibinfo{pages}{15501--15511}.
\bibitem[{Liu et~al.(2016)Liu, Zhang, Xue, Xu and Gao}]{liu2016registration}
\bibinfo{author}{Liu, Y.}, \bibinfo{author}{Zhang, H.}, \bibinfo{author}{Xue,
  Y.}, \bibinfo{author}{Xu, G.}, \bibinfo{author}{Gao, Z.},
  \bibinfo{year}{2016}.
\newblock \bibinfo{title}{Registration of depth maps based on irls-icp-tps},
  in: \bibinfo{booktitle}{2016 9th International Congress on Image and Signal
  Processing, BioMedical Engineering and Informatics (CISP-BMEI)},
  \bibinfo{organization}{IEEE}. pp. \bibinfo{pages}{543--547}.
\bibitem[{Ma et~al.(2016)Ma, Zhao and Yuille}]{7185406}
\bibinfo{author}{Ma, J.}, \bibinfo{author}{Zhao, J.}, \bibinfo{author}{Yuille,
  A.L.}, \bibinfo{year}{2016}.
\newblock \bibinfo{title}{Non-rigid point set registration by preserving global
  and local structures}.
\newblock \bibinfo{journal}{IEEE Transactions on Image Processing}
  \bibinfo{volume}{25}, \bibinfo{pages}{53--64}.
\bibitem[{Marin et~al.(2024)Marin, Corona and Pons-Moll}]{marin2024nicp}
\bibinfo{author}{Marin, R.}, \bibinfo{author}{Corona, E.},
  \bibinfo{author}{Pons-Moll, G.}, \bibinfo{year}{2024}.
\newblock \bibinfo{title}{Nicp: Neural icp for 3d human registration at scale},
  in: \bibinfo{booktitle}{Proc. of the European Conference on Computer Vision
  (ECCV)}, \bibinfo{organization}{Springer}. pp. \bibinfo{pages}{265--285}.
\bibitem[{Minar et~al.(2020)Minar, Tuan, Ahn, Rosin and Lai}]{minar2020cp}
\bibinfo{author}{Minar, M.R.}, \bibinfo{author}{Tuan, T.T.},
  \bibinfo{author}{Ahn, H.}, \bibinfo{author}{Rosin, P.}, \bibinfo{author}{Lai,
  Y.K.}, \bibinfo{year}{2020}.
\newblock \bibinfo{title}{Cp-vton+: Clothing shape and texture preserving
  image-based virtual try-on}, in: \bibinfo{booktitle}{{IEEE} Conf. on Computer
  Vision and Pattern Recognition Workshops (CVPRW)},
  \bibinfo{organization}{{IEEE}}. pp. \bibinfo{pages}{10--14}.
\bibitem[{Mirza and Osindero(2014)}]{mirza2014conditional}
\bibinfo{author}{Mirza, M.}, \bibinfo{author}{Osindero, S.},
  \bibinfo{year}{2014}.
\newblock \bibinfo{title}{Conditional generative adversarial nets}.
\newblock \bibinfo{journal}{arXiv preprint arXiv:1411.1784} .
\bibitem[{Myronenko and Song(2010)}]{5432191}
\bibinfo{author}{Myronenko, A.}, \bibinfo{author}{Song, X.},
  \bibinfo{year}{2010}.
\newblock \bibinfo{title}{Point set registration: Coherent point drift}.
\newblock \bibinfo{journal}{IEEE Transactions on Pattern Analysis and Machine
  Intelligence} \bibinfo{volume}{32}, \bibinfo{pages}{2262--2275}.
\bibitem[{Patel et~al.(2020)Patel, Liao and Pons-Moll}]{patel2020tailornet}
\bibinfo{author}{Patel, C.}, \bibinfo{author}{Liao, Z.},
  \bibinfo{author}{Pons-Moll, G.}, \bibinfo{year}{2020}.
\newblock \bibinfo{title}{Tailornet: Predicting clothing in 3d as a function of
  human pose, shape and garment style}, in: \bibinfo{booktitle}{{IEEE} Conf. on
  Computer Vision and Pattern Recognition (CVPR)},
  \bibinfo{organization}{{IEEE}}. pp. \bibinfo{pages}{7365–--7375}.
\bibitem[{Pons-Moll et~al.(2017)Pons-Moll, Pujades, Hu and
  Black}]{pons2017clothcap}
\bibinfo{author}{Pons-Moll, G.}, \bibinfo{author}{Pujades, S.},
  \bibinfo{author}{Hu, S.}, \bibinfo{author}{Black, M.J.},
  \bibinfo{year}{2017}.
\newblock \bibinfo{title}{Clothcap: Seamless 4d clothing capture and
  retargeting}.
\newblock \bibinfo{journal}{ACM Trans. on Graphics} \bibinfo{volume}{36},
  \bibinfo{pages}{1--15}.
\bibitem[{Salimans et~al.(2016)Salimans, Goodfellow, Zaremba, Cheung, Radford
  and Chen}]{salimans2016improved}
\bibinfo{author}{Salimans, T.}, \bibinfo{author}{Goodfellow, I.},
  \bibinfo{author}{Zaremba, W.}, \bibinfo{author}{Cheung, V.},
  \bibinfo{author}{Radford, A.}, \bibinfo{author}{Chen, X.},
  \bibinfo{year}{2016}.
\newblock \bibinfo{title}{Improved techniques for training gans}, in:
  \bibinfo{booktitle}{Advances in Neural Information Processing Systems
  (NIPS)}, \bibinfo{organization}{Springer}. pp. \bibinfo{pages}{2234--2242}.
\bibitem[{Schaefer et~al.(2006)Schaefer, McPhail and
  Warren}]{schaefer2006image}
\bibinfo{author}{Schaefer, S.}, \bibinfo{author}{McPhail, T.},
  \bibinfo{author}{Warren, J.}, \bibinfo{year}{2006}.
\newblock \bibinfo{title}{Image deformation using moving least squares}.
\newblock \bibinfo{journal}{ACM Trans. on Graphics} \bibinfo{volume}{25},
  \bibinfo{pages}{533--540}.
\bibitem[{Stein et~al.(2024)Stein, Cresswell, Hosseinzadeh, Sui, Ross,
  Villecroze, Liu, Caterini, Taylor and Loaiza-Ganem}]{stein2024exposing}
\bibinfo{author}{Stein, G.}, \bibinfo{author}{Cresswell, J.},
  \bibinfo{author}{Hosseinzadeh, R.}, \bibinfo{author}{Sui, Y.},
  \bibinfo{author}{Ross, B.}, \bibinfo{author}{Villecroze, V.},
  \bibinfo{author}{Liu, Z.}, \bibinfo{author}{Caterini, A.L.},
  \bibinfo{author}{Taylor, E.}, \bibinfo{author}{Loaiza-Ganem, G.},
  \bibinfo{year}{2024}.
\newblock \bibinfo{title}{Exposing flaws of generative model evaluation metrics
  and their unfair treatment of diffusion models}, in:
  \bibinfo{booktitle}{Advances in Neural Information Processing Systems
  (NIPS)}, \bibinfo{organization}{Springer}. pp. \bibinfo{pages}{3732--3784}.
\bibitem[{Takida et~al.(2024)Takida, Imaizumi, Shibuya, Lai, Uesaka, Murata and
  Mitsufuji}]{takida2024san}
\bibinfo{author}{Takida, Y.}, \bibinfo{author}{Imaizumi, M.},
  \bibinfo{author}{Shibuya, T.}, \bibinfo{author}{Lai, C.H.},
  \bibinfo{author}{Uesaka, T.}, \bibinfo{author}{Murata, N.},
  \bibinfo{author}{Mitsufuji, Y.}, \bibinfo{year}{2024}.
\newblock \bibinfo{title}{{SAN}: Inducing metrizability of {GAN} with
  discriminative normalized linear layer}, in: \bibinfo{booktitle}{The Twelfth
  International Conference on Learning Representations (ICLR)},
  \bibinfo{organization}{ICLR}. pp. \bibinfo{pages}{1--34}.
\bibitem[{Tulyakov et~al.(2018)Tulyakov, Liu, Yang and
  Kautz}]{tulyakov2018mocogan}
\bibinfo{author}{Tulyakov, S.}, \bibinfo{author}{Liu, M.Y.},
  \bibinfo{author}{Yang, X.}, \bibinfo{author}{Kautz, J.},
  \bibinfo{year}{2018}.
\newblock \bibinfo{title}{Mocogan: Decomposing motion and content for video
  generation}, in: \bibinfo{booktitle}{{IEEE} Conf. on Computer Vision and
  Pattern Recognition (CVPR)}, \bibinfo{organization}{{IEEE}}. pp.
  \bibinfo{pages}{1526--1535}.
\bibitem[{Usman~Akbar et~al.(2024)Usman~Akbar, Larsson, Blystad and
  Eklund}]{usman2024brain}
\bibinfo{author}{Usman~Akbar, M.}, \bibinfo{author}{Larsson, M.},
  \bibinfo{author}{Blystad, I.}, \bibinfo{author}{Eklund, A.},
  \bibinfo{year}{2024}.
\newblock \bibinfo{title}{Brain tumor segmentation using synthetic mr images-a
  comparison of gans and diffusion models}.
\newblock \bibinfo{journal}{Scientific Data} \bibinfo{volume}{11},
  \bibinfo{pages}{259--287}.
\bibitem[{Wang et~al.(2018a)Wang, Zheng, Liang, Chen, Lin and
  Yang}]{wang2018toward}
\bibinfo{author}{Wang, B.}, \bibinfo{author}{Zheng, H.},
  \bibinfo{author}{Liang, X.}, \bibinfo{author}{Chen, Y.},
  \bibinfo{author}{Lin, L.}, \bibinfo{author}{Yang, M.}, \bibinfo{year}{2018}a.
\newblock \bibinfo{title}{Toward characteristic-preserving image-based virtual
  try-on network}, in: \bibinfo{booktitle}{Proc. of the European Conference on
  Computer Vision (ECCV)}, \bibinfo{organization}{Springer}. pp.
  \bibinfo{pages}{589--604}.
\bibitem[{Wang et~al.(2023)Wang, Zhong, Sun, Chen, Zhao, Li and
  Wang}]{wang2023simultaneous}
\bibinfo{author}{Wang, H.}, \bibinfo{author}{Zhong, G.}, \bibinfo{author}{Sun,
  J.}, \bibinfo{author}{Chen, Y.}, \bibinfo{author}{Zhao, Y.},
  \bibinfo{author}{Li, S.}, \bibinfo{author}{Wang, D.}, \bibinfo{year}{2023}.
\newblock \bibinfo{title}{Simultaneous restoration and super-resolution gan for
  underwater image enhancement}.
\newblock \bibinfo{journal}{Frontiers in Marine Science} \bibinfo{volume}{10},
  \bibinfo{pages}{1162295}.
\bibitem[{Wang et~al.(2018b)Wang, Liu, Zhu, Tao, Kautz and
  Catanzaro}]{wang2018high}
\bibinfo{author}{Wang, T.C.}, \bibinfo{author}{Liu, M.Y.},
  \bibinfo{author}{Zhu, J.Y.}, \bibinfo{author}{Tao, A.},
  \bibinfo{author}{Kautz, J.}, \bibinfo{author}{Catanzaro, B.},
  \bibinfo{year}{2018}b.
\newblock \bibinfo{title}{High-resolution image synthesis and semantic
  manipulation with conditional gans}, in: \bibinfo{booktitle}{{IEEE} Conf. on
  Computer Vision and Pattern Recognition (CVPR)},
  \bibinfo{organization}{{IEEE}}. pp. \bibinfo{pages}{8798--8807}.
\bibitem[{Wang et~al.(2004)Wang, Bovik, Sheikh and Simoncelli}]{wang2004image}
\bibinfo{author}{Wang, Z.}, \bibinfo{author}{Bovik, A.C.},
  \bibinfo{author}{Sheikh, H.R.}, \bibinfo{author}{Simoncelli, E.P.},
  \bibinfo{year}{2004}.
\newblock \bibinfo{title}{Image quality assessment: from error visibility to
  structural similarity}.
\newblock \bibinfo{journal}{IEEE Trans. on Image Processing (TIP)}
  \bibinfo{volume}{13}, \bibinfo{pages}{600--612}.
\bibitem[{Wang et~al.(2021)Wang, She and Ward}]{wang2021generative}
\bibinfo{author}{Wang, Z.}, \bibinfo{author}{She, Q.}, \bibinfo{author}{Ward,
  T.E.}, \bibinfo{year}{2021}.
\newblock \bibinfo{title}{Generative adversarial networks in computer vision: A
  survey and taxonomy}.
\newblock \bibinfo{journal}{ACM Computing Surveys (CSUR)} \bibinfo{volume}{54},
  \bibinfo{pages}{1--38}.
\bibitem[{Yang et~al.(2020)Yang, Zhang, Guo, Liu and Zuo}]{yang2020towards}
\bibinfo{author}{Yang, H.}, \bibinfo{author}{Zhang, R.}, \bibinfo{author}{Guo,
  X.}, \bibinfo{author}{Liu, W.}, \bibinfo{author}{Zuo, W.},
  \bibinfo{year}{2020}.
\newblock \bibinfo{title}{Towards photo-realistic virtual try-on by adaptively
  generating-preserving image content}, in: \bibinfo{booktitle}{{IEEE} Conf. on
  Computer Vision and Pattern Recognition (CVPR)},
  \bibinfo{organization}{{IEEE}}. pp. \bibinfo{pages}{7850--7859}.
\bibitem[{Yang et~al.(2015)Yang, Ong and Foong}]{YANG2015156}
\bibinfo{author}{Yang, Y.}, \bibinfo{author}{Ong, S.H.},
  \bibinfo{author}{Foong, K.W.C.}, \bibinfo{year}{2015}.
\newblock \bibinfo{title}{A robust global and local mixture distance based
  non-rigid point set registration}.
\newblock \bibinfo{journal}{Pattern Recognition} \bibinfo{volume}{48},
  \bibinfo{pages}{156--173}.
\bibitem[{Yu et~al.(2019)Yu, Wang and Xie}]{yu2019vtnfp}
\bibinfo{author}{Yu, R.}, \bibinfo{author}{Wang, X.}, \bibinfo{author}{Xie,
  X.}, \bibinfo{year}{2019}.
\newblock \bibinfo{title}{Vtnfp: An image-based virtual try-on network with
  body and clothing feature preservation}, in: \bibinfo{booktitle}{{IEEE}
  International Conference on Computer Vision ({ICCV})},
  \bibinfo{organization}{{IEEE}}. pp. \bibinfo{pages}{10511--10520}.
\bibitem[{Zhang et~al.(2024)Zhang, Zhan, Sun, Zhang, Wei and
  Wang}]{zhang2024gic}
\bibinfo{author}{Zhang, P.}, \bibinfo{author}{Zhan, J.}, \bibinfo{author}{Sun,
  K.}, \bibinfo{author}{Zhang, J.}, \bibinfo{author}{Wei, M.},
  \bibinfo{author}{Wang, K.}, \bibinfo{year}{2024}.
\newblock \bibinfo{title}{Gic-flow: Appearance flow estimation via global
  information correlation for virtual try-on under large deformation}.
\newblock \bibinfo{journal}{Computers \& Graphics} \bibinfo{volume}{124},
  \bibinfo{pages}{104071}.
\bibitem[{Zhang et~al.(2018)Zhang, Isola, Efros, Shechtman and
  Wang}]{zhang2018unreasonable}
\bibinfo{author}{Zhang, R.}, \bibinfo{author}{Isola, P.},
  \bibinfo{author}{Efros, A.A.}, \bibinfo{author}{Shechtman, E.},
  \bibinfo{author}{Wang, O.}, \bibinfo{year}{2018}.
\newblock \bibinfo{title}{The unreasonable effectiveness of deep features as a
  perceptual metric}, in: \bibinfo{booktitle}{{IEEE} Conf. on Computer Vision
  and Pattern Recognition (CVPR)}, \bibinfo{organization}{{IEEE}}. pp.
  \bibinfo{pages}{586--595}.
\bibitem[{Zhou et~al.(2023)Zhou, Chen, Xiao and Huang}]{zhou2023neural}
\bibinfo{author}{Zhou, Y.}, \bibinfo{author}{Chen, K.}, \bibinfo{author}{Xiao,
  R.}, \bibinfo{author}{Huang, H.}, \bibinfo{year}{2023}.
\newblock \bibinfo{title}{Neural texture synthesis with guided correspondence},
  in: \bibinfo{booktitle}{{IEEE} Conf. on Computer Vision and Pattern
  Recognition (CVPR)}, \bibinfo{organization}{{IEEE}}. pp.
  \bibinfo{pages}{18095--18104}.
\bibitem[{Zhu et~al.(2017)Zhu, Park, Isola and Efros}]{zhu2017unpaired}
\bibinfo{author}{Zhu, J.Y.}, \bibinfo{author}{Park, T.},
  \bibinfo{author}{Isola, P.}, \bibinfo{author}{Efros, A.A.},
  \bibinfo{year}{2017}.
\newblock \bibinfo{title}{Unpaired image-to-image translation using
  cycle-consistent adversarial networks}, in: \bibinfo{booktitle}{{IEEE}
  International Conference on Computer Vision (ICCV)},
  \bibinfo{organization}{{IEEE}}. pp. \bibinfo{pages}{2223--2232}.
\bibitem[{Zhu et~al.(2020)Zhu, Xu, You and Bai}]{zhu2020semantically}
\bibinfo{author}{Zhu, Z.}, \bibinfo{author}{Xu, Z.}, \bibinfo{author}{You, A.},
  \bibinfo{author}{Bai, X.}, \bibinfo{year}{2020}.
\newblock \bibinfo{title}{Semantically multi-modal image synthesis}, in:
  \bibinfo{booktitle}{{IEEE} Conf. on Computer Vision and Pattern Recognition
  (CVPR)}, \bibinfo{organization}{{IEEE}}. pp. \bibinfo{pages}{5467--5476}.

\end{thebibliography}

\end{document}